\documentclass[sigconf]{acmart}
\renewcommand\footnotetextcopyrightpermission[1]{}
\AtBeginDocument{%
  \providecommand\BibTeX{{%
    \normalfont B\kern-0.5em{\scshape i\kern-0.25em b}\kern-0.8em\TeX}}}

\setcopyright{acmcopyright}
\copyrightyear{2018}
\acmYear{2018}
\acmDOI{XXXXXXX.XXXXXXX}

\acmPrice{15.00}
\acmISBN{978-1-4503-XXXX-X/18/06}

\usepackage{lipsum}
\usepackage{framed}
\usepackage{xcolor}
\usepackage{listings}
\hypersetup{
	colorlinks=true,
	urlcolor=magenta,
}

\usepackage{multirow}
\usepackage{rotating}
\usepackage{booktabs}
\usepackage{makecell}

\usepackage{mdframed}
\usepackage{amsmath}
\usepackage{colortbl}
\usepackage{algpseudocode}
\usepackage{algorithm} 
\usepackage{adjustbox}
\usepackage{color}
\usepackage{amsfonts}
\usepackage{bbding}
\definecolor{mygreen}{RGB}{149, 236, 105}
\newcommand{\algcom}[1]{\textcolor{mygreen}{\emph{// #1}}}
\begin{document}

\title{SUR-adapter: Enhancing Text-to-Image Pre-trained Diffusion Models with Large Language Models}
\author{Shanshan Zhong$^*$}
\email{zhongshsh5@mail2.sysu.edu.cn}
\orcid{0000-0003-3082-7351}
\affiliation{%
  \institution{Sun Yat-sen University}
  \city{Guangzhou}
  \country{China}
}

\author{Zhongzhan Huang}
\authornote{Both authors contributed equally to this research.}
\email{huangzhzh23@mail2.sysu.edu.cn}
\affiliation{%
  \institution{Sun Yat-sen University}
  \city{Guangzhou}
  \country{China}
}

\author{Wushao Wen}
\email{wenwsh@mail.sysu.edu.cn}
\affiliation{%
  \institution{Sun Yat-sen University}
  \city{Guangzhou}
  \country{China}
}

\author{Jinghui Qin}
\authornote{Corresponding author. }
\email{scape1989@gmail.com}
\affiliation{%
  \institution{Guangdong University of Technology}
  \city{Guangzhou}
  \country{China}
}

\author{Liang Lin}
\email{linliang@ieee.org}
\affiliation{%
  \institution{Sun Yat-sen University}
  \city{Guangzhou}
  \country{China}
}

\renewcommand{\shortauthors}{Shanshan Zhong, Zhongzhan Huang, WushaoWen, Jinghui Qin, \& Liang Lin}

\begin{abstract}
Diffusion models, which have emerged to become popular text-to-image generation models, can produce high-quality and content-rich images guided by textual prompts. However, there are limitations to semantic understanding and commonsense reasoning in existing models when the input prompts are concise narrative, resulting in low-quality image generation. To improve the capacities for narrative prompts, we propose a simple-yet-effective parameter-efficient fine-tuning approach called the Semantic Understanding and Reasoning adapter (SUR-adapter) for pre-trained diffusion models. To reach this goal, we first collect and annotate a new dataset SURD which consists of more than 57,000 semantically corrected multi-modal samples. Each sample contains a simple narrative prompt, a complex keyword-based prompt, and a high-quality image. Then, we align the semantic representation of narrative prompts to the complex prompts and transfer knowledge of large language models (LLMs) to our SUR-adapter via knowledge distillation so that it can acquire the powerful semantic understanding and reasoning capabilities to build a high-quality textual semantic representation for text-to-image generation. We conduct experiments by integrating multiple LLMs and popular pre-trained diffusion models to show the effectiveness of our approach in enabling diffusion models to understand and reason concise natural language without image quality degradation. Our approach can make text-to-image diffusion models easier to use with better user experience, which demonstrates our approach has the potential for further advancing the development of user-friendly text-to-image generation models by bridging the semantic gap between simple narrative prompts and complex keyword-based prompts. The code is released at \textcolor{magenta}{\url{https://github.com/Qrange-group/SUR-adapter}}.
\vspace{-0.3cm}
\end{abstract}

\begin{CCSXML}
<ccs2012>
   <concept>
       <concept_id>10010147.10010178.10010179</concept_id>
       <concept_desc>Computing methodologies~Natural language processing</concept_desc>
       <concept_significance>500</concept_significance>
       </concept>
   <concept>
       <concept_id>10010147.10010178.10010224</concept_id>
       <concept_desc>Computing methodologies~Computer vision</concept_desc>
       <concept_significance>500</concept_significance>
       </concept>
   <concept>
       <concept_id>10010147.10010257.10010321</concept_id>
       <concept_desc>Computing methodologies~Machine learning algorithms</concept_desc>
       <concept_significance>500</concept_significance>
       </concept>
 </ccs2012>
\vspace{-0.3cm}
\end{CCSXML}





\begin{teaserfigure}
  \includegraphics[width=\textwidth]{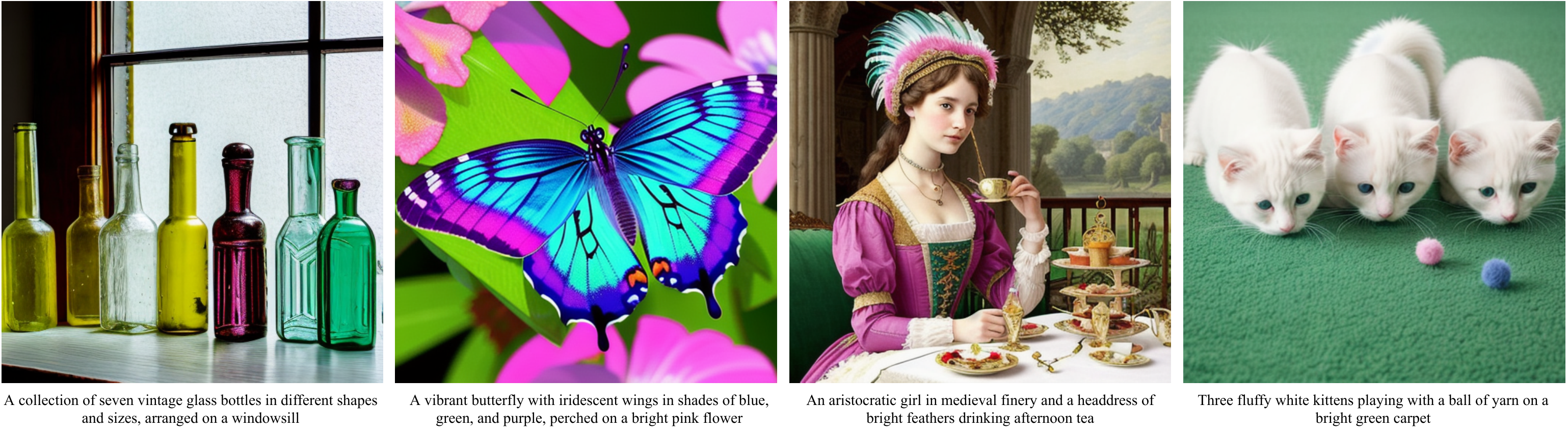}
  \caption{512×512 samples with various types of prompts (Counting, Color, Action, etc.), showing that SUR-adapter has powerful capabilities of fine-grained semantic control. }
  \label{fig:teaser}
\end{teaserfigure}

\maketitle

\vspace{-0.3cm}
\section{Introduction}
\label{sec:intro}

\begin{figure*}
  \centering
  \includegraphics[width=0.8\linewidth]{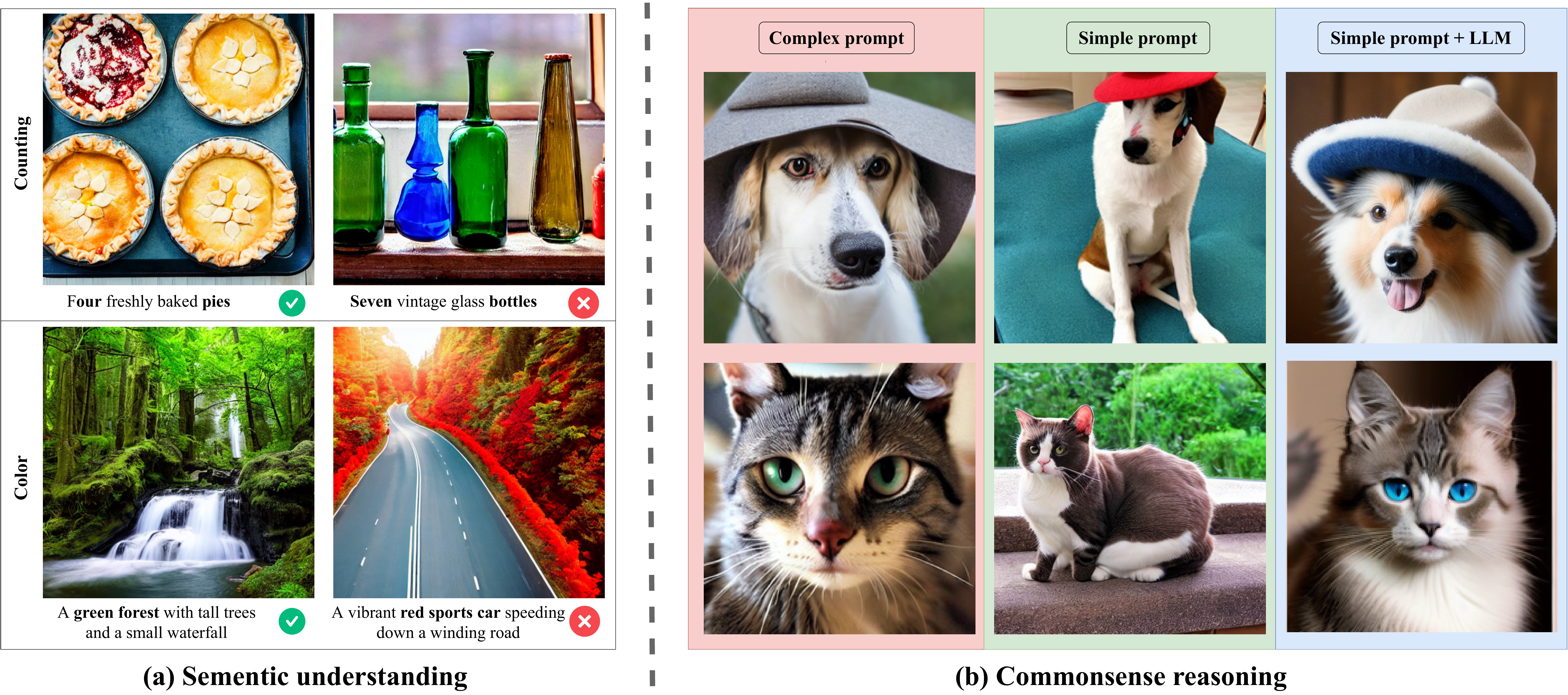}
  \caption{The semantic understanding and commonsense reasoning capability of text encoder in diffusion models.}
  \label{fig:intro}
  \vspace{-0.3cm}
\end{figure*}

In recent years, diffusion model based multimodal text-to-image generation techniques have made impressive strides \cite{Yang2022DiffusionMA}. With these models \cite{Rombach2021HighResolutionIS,Valevski2022UniTuneTI} trained on massive amounts of data and model parameters, people are able to generate text-relevant and visually appealing images end-to-end through text prompts and other information, without requiring complex painting skills. However, the quality of image generation in these existing diffusion models heavily relies on the complex and elaborate design of keyword-based text prompts or other forms of text prompts. Furthermore, if the text prompts are concise narratives or short phrases that are daily expressions, the fidelity and text relevance of the generated images are often significantly compromised. 
This limitation makes diffusion models difficult to be controlled intuitively by concise narratives with excellent user experience. 
The most important reason for this issue is that the text encoders of these diffusion models, which are often the text encoder of pre-trained CLIP \cite{radford2021learning} trained with image-text contrastive learning, are unaligned to text-to-image generation task, leading to poor semantic understanding and reasoning (SUR) for image generation.

To be specific, CLIP is a multi-modal neural model trained on about 400M image-text pairs with contrastive learning and its image encoder and text encoder have been widely applied in various multi-modal tasks or models, such as diffusion models, since it can bridge the association between images and text successfully. Although the learning objective of CLIP is to establish image-text correspondence by only pulling the matched image and text pair closer in feature space, the text describing the corresponding image is brief and may only match partial semantics in the image, resulting in the incomplete feature generated by the text encoder. However, the text-to-image generation task asks a text encoder can not only understand the semantics of a concise narrative but also reason out and complete the implicit commonsense or knowledge grounded in the narrative so that a model can generate an accurate image that is highly consistent with the narrative. Therefore, embedding the text encoder of CLIP into diffusion models for conditional text-to-image generation results in low-quality image generation when the input text is natural language due to a lack of the capability of semantic understanding and commonsense reasoning in the text encoder.

\begin{table}[t]
  \centering
  \caption{Evaluation of semantic accuracy (Acc.) in images generated by simple prompts using diffusion models. The simple prompts consisted of three types of sentences, including "Counting", "Color", and "Action". Each prompt generated 130 images, and the images were manually checked for semantic accuracy. The results showed that the semantic accuracy of most prompts is below 50\%, and even two types of prompts have the semantic accuracy rate of 0\%. }
  \resizebox*{\linewidth}{!}{
    \begin{tabular}{llrc}
    \toprule
    Type & Prompt & Acc. & $<$ 50\%?\\
    \midrule
    \multirow{3}[2]{*}{Counting} & Four freshly baked pies. & 63.08\%  &\\
          & Six colorful balloons floating over a picturesque landscape. & 8.46\% &\checkmark\\
          & Seven vintage glass bottles. & {\color{red}\textbf{0.00}\%} &\checkmark\\
    \midrule
    \multirow{3}[2]{*}{Color} & A vibrant red sports car speeding down a winding road. & 86.15\% & \\
          & The blue glass containing red juice.  & 17.69\% &\checkmark\\
          & A couple wearing blue and yellow solid color clothes. & {\color{red}\textbf{0.00}\%} &\checkmark\\
    \midrule
    \multirow{3}[2]{*}{Action} & Someone shooting a basketball on the sports field. & 41.54\% &\checkmark\\
          & Giraffes eating trees. & 25.38\% &\checkmark\\
          & A chef tossing a pizza dough in the air in a kitchen. & 15.38\% &\checkmark\\
    \bottomrule
    \end{tabular}%
  }
  \label{tab:promptacc}%
\vspace{-0.5cm}
\end{table}%

To show these deficiencies, we first evaluate the semantic understanding capability of the text encoder in diffusion models using three common types of text prompts in multi-modal visual question answering \cite{Chen2022RethinkingDA,Chen2020CounterfactualSS,Agrawal2015VQAVQ,Liu2021SlakeAS}: "counting", "color", and "action".
As shown in Table \ref{tab:promptacc}, we designed three different prompts for each type, and for each text prompt, we generated 130 images using a text-to-image diffusion model \cite{Rombach2021HighResolutionIS} and manually evaluated whether the generated images fulfilled the semantic meaning of the given text prompts. Through statistical analysis of the results, we found that the accuracy of semantic understanding for most of the text prompts does not exceed 50\%.  Surprisingly, even seemingly simple narrative prompts such as "Seven vintage glass bottles" and "A couple wearing blue and yellow solid color clothes respectively" have 0\% accuracy, indicating that the text encoder in the diffusion model completely fails to understand the semantics of these simple texts for image generation and resulted in severe information bias. Fig.\ref{fig:intro}(a) further illustrates examples of semantic error due to inadequate semantic understanding capability.

Next, we consider the commonsense reasoning ability of the text encoder. If we hope the stable diffusion model to generate a beautiful cat, according to widely verified generation techniques, we need some complex and elaborate keyword-based prompts to obtain high-quality generated images, such as the following prompt:
\vspace{0.05cm}
\begin{mdframed}[backgroundcolor=gray!8]
\begin{minipage}{\linewidth}
(\textbf{Complex prompt example}) 8k uhd, a RAW photo, a beautiful cat, (realistic:1.1), masterwork, RAW photo, real cat, RAW photograph, ultra high res, photorealistic, best quality, (high detailed skin,skin details), visible pores, shiny skin, an extremely delicate and beautiful, extremely detailed 8K wallpaper, 8k high quality, film grain, DSLR, beautiful cat with beautiful details, (looking at viewer), professional photography lighting, extremely detailed eyes and face, eyes with beautiful details, analog style, cute and playful, adorable, (splendid and colorful:1.1), portrait picture of cat, <lora:mikeneko:0.7>, from side, full body, (brown black white fur)
\end{minipage}
\end{mdframed}
\vspace{0.1cm}

We can observe that images generated using complex prompts, as shown in Fig.\ref{fig:intro} (b), have better details, more accurate outlines, and precise common sense (such as the cat's body is natural and non-distorted) compared to images generated using simple prompts like "a beautiful cat".
\vspace{0.1cm}
\begin{mdframed}[backgroundcolor=gray!8]
\begin{minipage}{\linewidth}
(\textbf{Simple prompt example}) a beautiful cat
\end{minipage}
\end{mdframed}
\vspace{0.1cm}

Inputting complex prompts is equivalent to directly injecting the details and understanding between "beautiful" and "cat" into the text encoder, allowing diffusion models to generate a pleasing "beautiful cat". This indicates that diffusion models have the potential to generate semantically meaningful images, but are limited by the text encoder's commonsense reasoning ability. Simple prompts do not allow the text encoder to directly understand the meaning of "beautiful cat" well, nor can it deduce the meaning of "beautiful" from the encoder's own knowledge. Facing such a problem, recent advances in large language models (LLMs) such as ChatGPT and LLaMA \cite{touvron2023llama} have shown astonishing conversational capabilities, with improved SUR abilities, creating new heights in the field of natural language processing (NLP). Therefore, we made an attempt to describe "a beautiful cat" using ChatGPT and obtained the following text: 
\vspace{0.1cm}
\begin{mdframed}[backgroundcolor=gray!8]
\vspace{0.1cm}
\begin{minipage}{\linewidth}
(\textbf{Commonsense reasoning of LLM}) Cats are known for their captivating beauty, with their soft fur,  expressive eyes, and graceful movements. A beautiful cat might have distinctive features such as a sleek coat with unique patterns, piercing  eyes, and an elegant posture. Each cat is unique in its own way, and  their beauty is subjective to the beholder's perspective.
\end{minipage}
\end{mdframed}
\vspace{0.1cm}
This text demonstrates the LLMs understanding of "beautiful" and "cat", as well as its commonsense reasoning on what kind of "cat" is considered "beautiful". The image produced by this text is similar in quality to images generated using complex prompts, as shown in Fig.\ref{fig:intro} (b) bottom right.

All of the case studies above inspire us to consider whether we can transfer the SUR abilities of LLMs to pre-trained diffusion models so that diffusion models can produce semantically correct and high-quality images even with simple narrative prompts.

To achieve this goal, in this paper, we first collect and annotate a new dataset named SURD, which consists of more than 57,000 semantically corrected image-text pairs. Each image-text pair contains a simple narrative prompt, a complex keyword-based prompt, and a high-quality image. Leveraging SURD, we propose the SUR-adapter to transfer the SUR abilities of LLMs to pre-trained diffusion models and align the representations of simple and complex prompts. Extensive experiments and statistical tests confirm that our proposed SUR-adapter significantly enhances the text encoder of pre-trained diffusion models and generates high-quality images that alleviate the mismatch between concise narrative prompts and generated images. In summary, our contributions are threefold:

\begin{itemize}

    \item We collect and annotate a dataset SURD, which includes over 57,000 semantically corrected image-text pairs. Each image-text pair contains a simple prompt, a complex prompt, and a high-quality corresponding image. 
    \item Based on SURD, we propose SUR-adapter to effectively transfer the semantic understanding and reasoning abilities of LLMs to pre-trained diffusion models, alleviating the issue of semantic mismatch and low-quality images generated with simple prompts.
    \item We conduct extensive statistical tests and discussions on the generated images using the proposed SUR-adapter to analyze its effectiveness and further discuss its limitations.
\end{itemize}

\section{Related works}

\subsection{Text-to-Image Diffusion}

Diffusion models have been extensively utilized in text-to-image generation \cite{Rombach2021HighResolutionIS,Bao2023OneTF,Kim2021DiffusionCLIPTD,Valevski2022UniTuneTI,Kawar2022ImagicTR,Ruiz2022DreamBoothFT,Fan2022FridoFP}. Text-to-image diffusion utilizes textual input as a conditioning signal for diffusion models, generating text-related images via a process of noise addition and removal~\cite{Rombach2021HighResolutionIS}. The text encoder of text-to-image diffusion is often accomplished by leveraging pre-trained language models such as CLIP \cite{radford2021learning} to encode textual inputs into latent vectors. Text-to-image diffusion is widely used in various fields, such as image super-resolution \cite{Li2021SRDiffSI,Saharia2021ImageSV}, inpainting \cite{Lugmayr2022RePaintIU}, manipulation \cite{Batzolis2021ConditionalIG,Zhang2023AddingCC}, semantic segmentation \cite{Baranchuk2021LabelEfficientSS,Graikos2022DiffusionMA}, video generation \cite{Zhang2022MotionDiffuseTH,Yang2022DiffusionPM}, etc. 

\subsection{Large Language Models}
Recently, the NLP field has witnessed a proliferation of LLMs \cite{Hoffmann2022TrainingCL}. Jozefowicz et al. \cite{2016Exploring} achieved state-of-the-art results on the Billion Word benchmark by scaling LSTMs to 1 billion parameters. Subsequently, scaling transformers led to improvements on many NLP tasks, with notable models including BERT \cite{2018BERT}, GPT-2 \cite{Radford2019LanguageMA}, MegatronLM \cite{Shoeybi2019MegatronLMTM}, and T5 \cite{Raffel2019ExploringTL}. The introduction of GPT-3 \cite{Brown2020LanguageMA}, a model with 175 billion parameters, marked a significant breakthrough in this area and led to the development of numerous LLMs, such as Jurassic-1 \cite{lieber2021jurassic}, Megatron-Turing NLG \cite{Smith2022UsingDA}, Gopher \cite{Rae2021ScalingLM}, Chinchilla \cite{Hoffmann2022TrainingCL}, PaLM \cite{Chowdhery2022PaLMSL}, OPT \cite{Zhang2022OPTOP}, GLM \cite{Zeng2022GLM130BAO} and LLAMA \cite{touvron2023llama}. 
Furthermore, several studies \cite{Hestness2017DeepLS,Rosenfeld2019ACP,Kaplan2020ScalingLF,Hoffmann2022TrainingCL,Wei2022EmergentAO} have investigated the impact of scaling on LLM performance to enhance their ease of use.

\section{Semantic Understanding and Reasoning Dataset}
\label{sec:datasetsurd}

\begin{figure}
  \centering
  \includegraphics[width=0.9\linewidth]{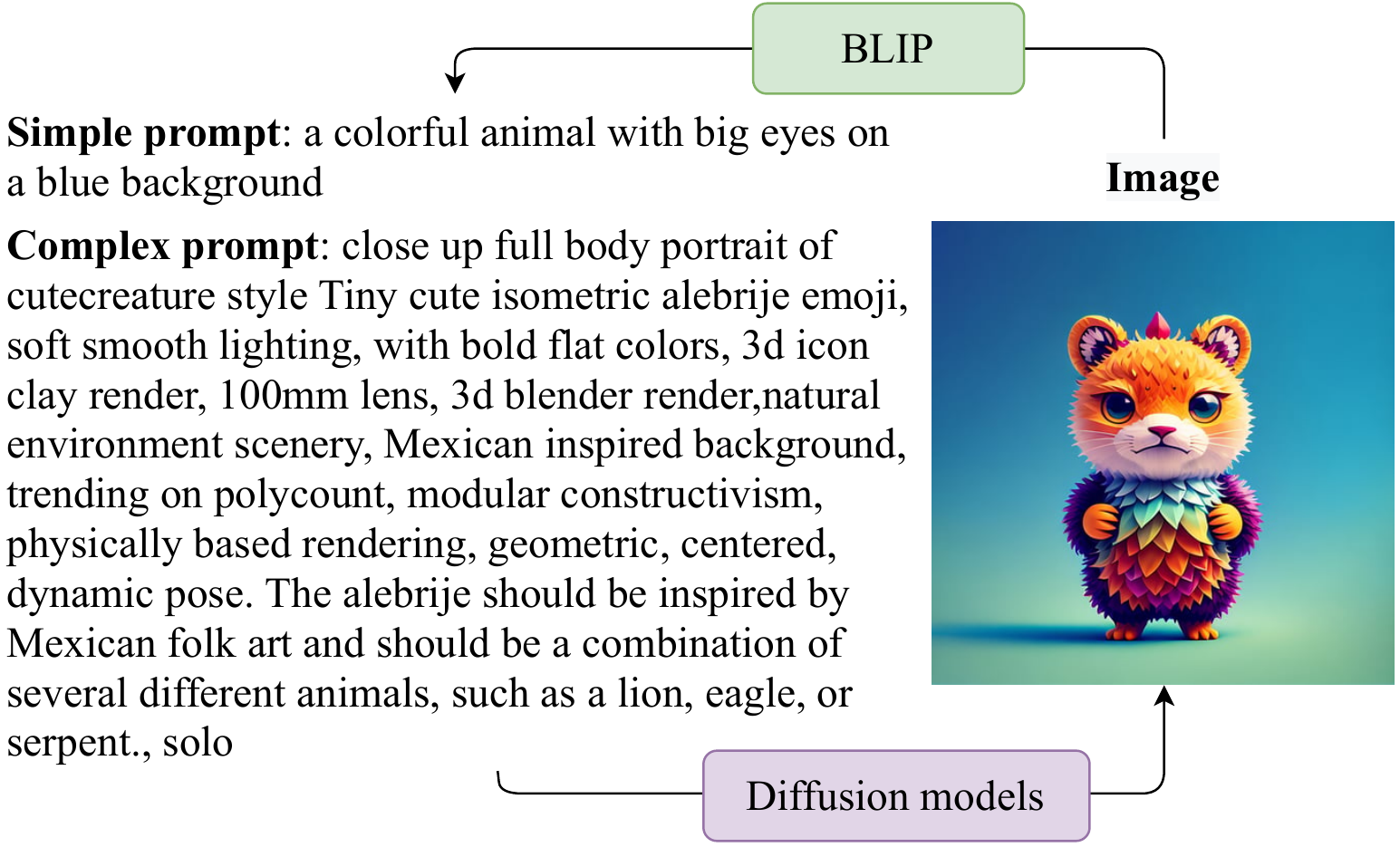}
  \caption{An example of SURD. We collect a diverse set of complex prompts and corresponding images generated by diffusion models from publicly available websites and leverage pre-trained BLIP to generate simple prompts. }
  \label{fig:dataset_example}
\end{figure}

\begin{figure*}
  \centering
  \includegraphics[width=0.8\linewidth]{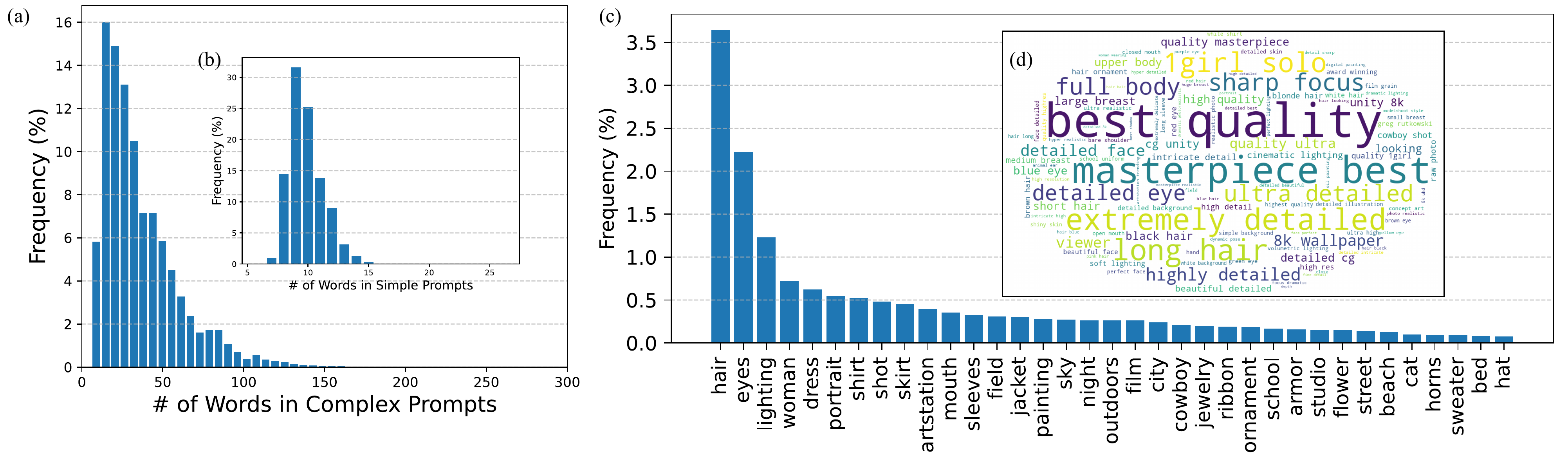}
  \caption{(Left) Prompt length distributions and (Right) prompt content distributions.}
  \label{fig:data}
\end{figure*}

SURD is a multi-modal dataset comprising 57,603 triplets of simple narrative prompts, complex keyword-based prompts, and semantically correct images, as shown in Fig. \ref{fig:dataset_example}. To our knowledge, SURD is the first dataset that records both simple and complex prompts and focuses on providing semantically correct image-text pairs to aid in solving the SUR problem of text-to-image diffusion models, which allows diffusion models to generate high-quality images that are semantically consistent based on simple prompts alone. 

\subsection{Data Collection}
\textbf{Raw Data.} To construct a content-rich and semantically reliable dataset, we extensively investigate various open-source image generation websites with reliable prompts and high-quality images. Among them, we select three websites: Lexica \footnote{https://lexica.art, $^2$https://civitai.com, $^3$https://stablediffusionweb.com}, civitai $^2$, and Stable Diffusion Online $^3$. On these websites, publicly available images are often semantically correct and of high quality with complex prompts. Therefore, we collect the prompts from websites as complex prompts. In total, we collect 114,148 image-text pairs. 

\noindent\textbf{Data Cleaning}. In order to ensure the correct semantic match of each sample in the SURD, we perform data cleaning in two steps. In the first step, to ensure the semantic accuracy of the simple prompts generated by BLIP \cite{guo2022images}, we use the publicly available pre-trained model CLIP \cite{radford2021learning} for semantic cleaning since the text encoder in most of diffusion models is the text encoder of the CLIP model, which will be explained in Section \ref{sec:preliminary}. If the CLIP model judges the semantic of a simple prompt that matches the semantic of the corresponding image, diffusion models are likely to be able to generate similar images according to the simple prompt. For each image, we ask CLIP to classify between its simple prompt and its complex prompt for selecting a prompt matching the image best semantically. In general, a complex prompt often contains other semantically irrelevant information, such as image quality descriptions, so a semantically correct simple prompt generally has a higher CLIP score than the complex prompt. Therefore, if the CLIP score of a simple prompt is not lower than the corresponding complex prompt, we retain the sample. After the automatic semantic cleaning based on the CLIP score, we retain 66,408 samples. In the second step, we further filter the samples retained in the first step manually to ensure that all image-text pairs are semantically matched. Finally, SURD contains 57,603 image-text pairs where each image-text pair contains an image, a simple prompt, and a complex prompt.  

\noindent\textbf{Knowledge from LLM. } Since we hope to distill knowledge from LLM to improve the semantic understanding and reasoning capacities of a text encoder, we also save the knowledge of simple prompts from LLM in vectors. Specifically, we use the recently open-sourced large language model LLaMA \cite{touvron2023llama} with three different parameter sizes: 7B (32 layers, dimension is 4096), 13B (40 layers, dimension is 5120), and 33B (60 layers, dimension is 6656). For each simple prompt, we compute the mean value of each token embeddings generated by the LLM as the knowledge representation so that we can handle different samples with different lengths uniformly. 

In addition, we resize all images to 512 $\times$ 512 uniformly. Further details regarding the usage of BLIP, CLIP, and LLM can be found in the appendix.

\subsection{Data Analysis}
\label{sec:data_ana}

\noindent\textbf{Prompt Length.} Fig.~\ref{fig:data} shows the distribution of sentence length for prompts, with (a) representing the distribution for complex prompts and (b) representing the distribution for simple prompts. In order to enhance visual clarity, prompts longer than 300 words have been incorporated into 300 words. The length distribution of simple prompts is relatively concentrated, with sentence lengths centered around 10, which is consistent with human language patterns. In contrast, complex prompts, with a long tail distribution, not only contain semantics but also include definitions and image quality information, resulting in sentence lengths that are significantly longer than simple prompts.

\noindent\textbf{Prompt Content.} A prompt for text-to-image generation usually contains a significant number of nouns which could influence the quality and semantic coherence of the generated image greatly since an image consists of different objects. Therefore, we conduct a statistical analysis of the frequency distribution of nouns occurred in the SURD to demonstrate the diversity of both text and visual content. Fig.~\ref{fig:data} (c) displays the frequency-proportional distribution of selected entities from SURD. These entities cover a diverse range of ordinary objects, such as people, animals, plants, and scenes, indicating the content diversity of SURD. Besides, the diversity of these entities can make pre-trained diffusion models have strong high-level understanding capacities of text and visual content in more complex scenes. Furthermore, we also present a word cloud of the text as shown in Fig. \ref{fig:data} (d) by filtering out stop words to illustrate the overall distribution of text vocabulary in SURD. The most frequently occurred phrases, such as "best quality", "masterpiece best", and "extremely detailed", primarily constrain the image quality and originate from complex prompts, indicating that these consistent text constraints are important for high-quality image generation. Therefore, the semantic representation of complex prompts will play a crucial role in enhancing the diffusion models with SUR-adapter through finetuning.

\begin{figure}
  \centering
  \includegraphics[width=0.99\linewidth]{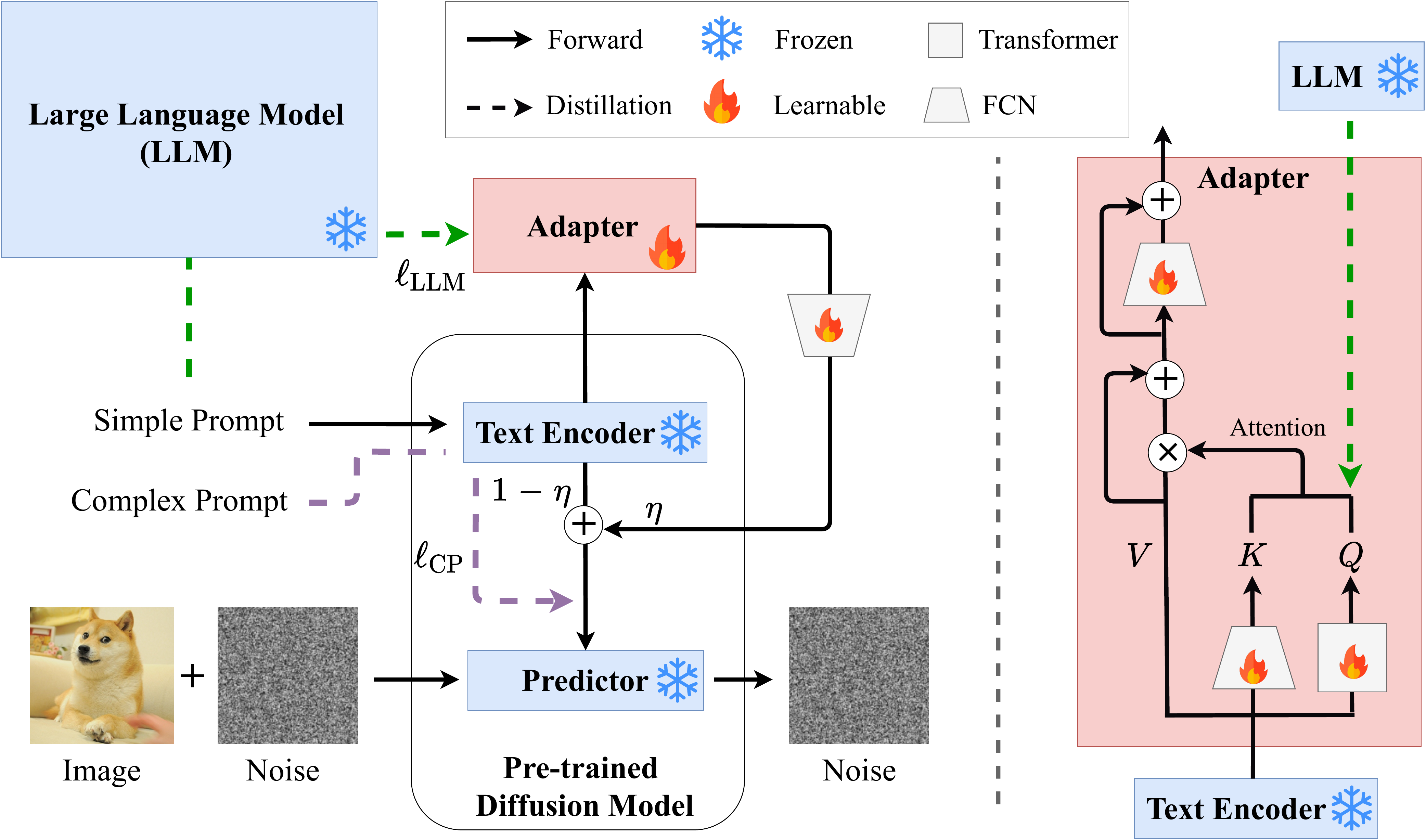}
  \caption{Illustration of SUR-adapter. FCN is a fully-connected network. (Left) The fine-tuning pipeline for pre-trained diffusion models. Given a pre-trained diffusion model, the adapter is used to transfer the semantic understanding and reasoning capabilities of large language models and align the representation between complex and simple prompts. The weight coefficient, $\eta$, is used to adjust the adapter's effect. (Right) The network structure of the adapter. }
  \label{fig:jg}
  \vspace{-0.2cm}
\end{figure}

\section{Method}
\label{sec:method}
In this section, we introduce how SUR-adapter transfers the semantic understanding and reasoning capabilities of large language models and achieves the representation alignment between complex prompts and simple prompts.

\subsection{Preliminary}
\label{sec:preliminary}

Diffusion models are excellent methods for multi-modal image generation, which typically involve two stages: \textbf{(1) Forward noise process.} Assuming that the training data $\mathbf{x}_0$ comes from a given distribution $p(\mathbf{x}_0)$, the diffusion model first obtains a sequence ${\mathbf{x}_1,\mathbf{x}_2,...,\mathbf{x}_T}$ by adding $T$ rounds of noise to $\mathbf{x}_0$, as follows:
\begin{equation}
q(\mathbf{x}_t|\mathbf{x}_0) = N(\mathbf{x}_t; \alpha_t\mathbf{x}_0,\sigma_t^2\mathbf{I}),
\label{eq:xt}
\end{equation}
where $\epsilon$ is sampled from the standard normal distribution $N(\mathbf{0},\mathbf{I})$, $\sigma_t^2$ is a given noise strength that depends on $t$, and $\alpha_t$ is generally set to $\alpha_t = \sqrt{1-\sigma_t^2}$. At this point, we have $\mathbf{x}_t = \alpha_t\mathbf{x}_0+\sigma_t\epsilon$. \textbf{(2) Reverse denoising process.} After obtaining the sequence ${\mathbf{x}_1,\mathbf{x}_2,...,\mathbf{x}_T}$ from the forward noise process, the denoising process from $\mathbf{x}_t$ to $\mathbf{x}_{t-1}$ can be modeled by
\begin{equation}
p_\theta(\mathbf{x}_{t-1}|\mathbf{x}_t) = N(\mathbf{x}_{t-1};\hat{\mu}_\theta(\mathbf{x}_t),\hat{\Sigma}_\theta(\mathbf{x}_t)),
\label{eq:rever}
\end{equation}
where $\hat{\mu}_\theta(\mathbf{x}t)$ and $\hat{\Sigma}_\theta(\mathbf{x}t))$ are the predicted statistics, and $\theta$ is the learnable parameter. Many recent works~\cite{Zhang2023TexttoimageDM,Rombach2021HighResolutionIS,Ho2020DenoisingDP} have shown that Eq.(\ref{eq:rever}) can be efficiently optimized via the following loss function:
\begin{equation}
\ell_\text{simple}^t(\theta) = \mathbb{E} \| \epsilon - \hat{\epsilon}_\theta (\alpha_t\mathbf{x}_0+\sigma_t\epsilon, t)  \|_2^2 ,
\label{eq:loss_main}
\end{equation}
where $\hat{\epsilon}_\theta(\cdot)$ is a learnable neural network that predicts the added noise $\epsilon$ in the input $\mathbf{x}_t$. When this neural network is well-trained, we can use $\mathbf{x}_t = \alpha_t\mathbf{x}_0+\sigma_t\epsilon$ and some certain sampling methods to infer $\mathbf{x}_0$. Note that as $T\to\infty$ or becomes sufficiently large, $\mathbf{x}_T$ can be viewed as an approximation of a normally distributed noise. Therefore, we can randomly sample noise $\epsilon_0$ from a normal distribution and use the neural network $\hat{\epsilon}_\theta(\cdot)$, also known as the predictor (as shown in Fig. \ref{fig:jg}), to generate an image $\hat{\mathbf{x}}_0$. To achieve controllable generation, a condition $c$ can be added to the predictor, i.e., rewriting the predictor as $\hat{\epsilon}_\theta(\cdot, c)$. For text-to-image generation tasks, the condition $c$ is usually generated from a text prompt by a text encoder, such as the text encoder of CLIP.

\begin{table*}[htbp]
  \centering
  \caption{Evaluation results of the diverse pre-trained models and controlled methods described in Section \ref{sec:imp} in terms of various semantic metrics. }
  \resizebox*{0.7\linewidth}{!}{
    \begin{tabular}{llcrcrcrcr}
    \toprule
    \multicolumn{1}{l}{\multirow{2}[4]{*}{\makecell{Pre-trained\\Model}}} & \multirow{2}[4]{*}{Controlled Method} & \multicolumn{2}{c}{CLIP Score} & \multicolumn{2}{c}{Action (\%)} & \multicolumn{2}{c}{Color (\%)} & \multicolumn{2}{c}{Counting (\%)} \\
\cmidrule{3-10}          &       & Baseline & Ours  & Baseline & Ours  & Baseline & Ours  & Baseline & Ours \\
    \midrule
    \multirow{6}[2]{*}{\makecell{DM (1.5),\\LLM (13B)}} & -     & 0.498 & 0.517 {\color{red}{$\uparrow$}} & 75.33 & 80.67 {\color{red}{$\uparrow$}} & 81.33 & 87.33 {\color{red}{$\uparrow$}} & 14.67 & 36.67 {\color{red}{$\uparrow$}} \\
          & ControlNet (canny) & 0.508 & 0.492 {\color{green}{$\downarrow$}} & 76.67 & 84.67 {\color{red}{$\uparrow$}} & 68.67 & 69.33 {\color{red}{$\uparrow$}} & 96.00 & 94.00 {\color{green}{$\downarrow$}} \\
          & ControlNet (seg) & 0.481 & 0.472 {\color{green}{$\downarrow$}} & 7.33  & 9.33 {\color{red}{$\uparrow$}}  & 10.00 & 10.67 {\color{red}{$\uparrow$}}  & 40.67 & 62.00 {\color{red}{$\uparrow$}} \\
          & Prompt Weighting & 0.486 & 0.514 {\color{red}{$\uparrow$}} & 78.00 & 85.33 {\color{red}{$\uparrow$}} & 91.33 & 88.00 {\color{green}{$\downarrow$}} & 43.33 & 58.00 {\color{red}{$\uparrow$}} \\
          & MultiDiffusion & 0.470 & 0.516 {\color{red}{$\uparrow$}} & 74.67 & 88.67 {\color{red}{$\uparrow$}} & 87.33 & 81.33 {\color{green}{$\downarrow$}} & 23.33 & 62.67 {\color{red}{$\uparrow$}} \\
          & Self-attention Guidance & 0.474 & 0.526 {\color{red}{$\uparrow$}} & 73.33 & 86.00 {\color{red}{$\uparrow$}} & 86.00 & 86.67 {\color{red}{$\uparrow$}} & 12.67 & 14.00 {\color{red}{$\uparrow$}} \\
    \midrule
    \multirow{6}[2]{*}{\makecell{DM (cartoon),\\LLM (13B)}} & -     & 0.467 & 0.490 {\color{red}{$\uparrow$}} & 58.00 & 68.67 {\color{red}{$\uparrow$}} & 82.00 & 88.00 {\color{red}{$\uparrow$}} & 21.33 & 38.00 {\color{red}{$\uparrow$}} \\
          & ControlNet (canny) & 0.514 & 0.486 {\color{green}{$\downarrow$}} & 83.33 & 81.33 {\color{green}{$\downarrow$}} & 47.33 & 67.33 {\color{red}{$\uparrow$}} & 74.00 & 86.67 {\color{red}{$\uparrow$}} \\
          & ControlNet (seg) & 0.509 & 0.491 {\color{green}{$\downarrow$}} & 38.67 & 51.33 {\color{red}{$\uparrow$}} & 28.00 & 30.67 {\color{red}{$\uparrow$}} & 45.33 & 62.00 {\color{red}{$\uparrow$}} \\
          & Prompt Weighting & 0.554 & 0.546 {\color{green}{$\downarrow$}} & 84.00 & 79.33 {\color{green}{$\downarrow$}} & 88.67 & 91.33 {\color{red}{$\uparrow$}} & 41.33 & 50.00 {\color{red}{$\uparrow$}} \\
          & MultiDiffusion & 0.413 & 0.587 {\color{red}{$\uparrow$}} & 63.33 & 80.67 {\color{red}{$\uparrow$}} & 88.00 & 87.33 {\color{green}{$\downarrow$}} & 18.67 & 36.67 {\color{red}{$\uparrow$}} \\
          & Self-attention Guidance & 0.440 & 0.560 {\color{red}{$\uparrow$}} & 65.33 & 73.33 {\color{red}{$\uparrow$}} & 72.67 & 86.00 {\color{red}{$\uparrow$}} & 16.67 & 39.33 {\color{red}{$\uparrow$}} \\
    \bottomrule
    \end{tabular}%
    }
  \label{tab:semanteme}%
\end{table*}%

\begin{algorithm}
\small
  \caption{The Algorithm of Fine-tuning Pre-trained Diffusion Model with SUR-adapter.}
  \label{alg:ean}  
  \begin{algorithmic}[1]
    \State \textbf{Input:} The dataset SURD $(p_c^i,p_s^i,I_i)_{i=1}^N$, a learnable transformation $g(\cdot;\phi_1)$ and Adapter $g_{\text{Ada}(\cdot;\phi_2)}$; Large language model $f_{\text{LLM}}$ and the text encoder $f_{\text{En}}$ with fixed parameters. Training step $T_0$.

    \While{The training step $T_0 \geq 0$}
    \State\algcom{Knowledge distillation from LLM}
    \State Calculate the knowledge distillation loss $\ell_{\text{LLM}}$ by Eq.(\ref{eq:lossllm})
    \State Measure the semantic information $c_{\text{LLM}}^\prime$ by Eq.(\ref{eq:predinput})
    \State\algcom{Performance maintenance}
    \State Add noise to $I_i$ to obtain $\alpha_t I_i+\sigma_t\epsilon$ by Eq.(\ref{eq:xt})
    \State Use $c_{\text{LLM}}^\prime$ to measure $\ell_\text{simple}^t(\phi)$ by Eq.(\ref{eq:mianloss})
    \State\algcom{Representation alignment}
    \State Measure $f_{En}(p_c^i)$ by complex prompt $p_c^i$
    \State Use $c_{\text{LLM}}^\prime$ and $f_{En}(p_c^i)$ to measure $\ell_{\text{CP}}(\phi)$ by Eq.(\ref{eq:losscp})
\State\algcom{Update the parameters}
    
\State Calculate the total loss $\ell_{\text{total}}(\phi)$ by Eq.(\ref{eq:totalloss})
\State Update the learnable parameters $\phi = [\phi_1,\phi_2]$ by $\ell_{\text{total}}(\phi)$
\State $T_0 \gets T_0 -1 $
\EndWhile

    \State\Return $\phi$
    
  \end{algorithmic}
\end{algorithm}

\subsection{The Fine-tuning Pipeline of SUR-adapter}
\label{sec:sura}

In this section, we introduce our simple yet effective fine-tuning approach called the semantic understanding and reasoning adapter (SUR-adapter) for the controllable text-to-image diffusion model. Let us consider the image-text pairs ${(p_{c}^i,p_{s}^i,I^i)}_{i=1}^N$ in the SURD dataset, where $p_c^i$ and $p_{s}^i$ are the complex and simple prompts, respectively, for the $i$-th high-quality and semantically correct image $I_i$. As shown in Fig.\ref{fig:jg} (Left), we first freeze all learnable parameters of the large language model $f_{\text{LLM}}$, the text encoder $f_{En}$, and the predictor $f_{pre}$ in the pre-trained diffusion model, and then we add two trainable neural networks, a fully-connected network (FCN) $g(\cdot;\phi_1)$ and an adapter $g_{\text{Ada}}(\cdot;\phi_2)$, with learnable parameters $\phi_1$ and $\phi_2$. 
%$\phi = [\phi_1,\phi_2]$.

%\noindent\textbf{Knowledge Distillation by LLM}. 
\subsubsection{Knowledge Distillation by LLM}
The structure of the adapter $g_{\text{Ada}}(\cdot;\phi_2)$ is shown in Fig.~\ref{fig:jg} (Right), and it consists of three learnable transformations, $h_j(\cdot)$ for $j=1,2,3$, which are implemented using fully connected neural networks or Transformer \cite{vaswani2017attention}. For the output $f_{En}(p_s^i)$ of the text encoder, we construct $Q_i = h_3[f_{En}(p_s^i)]$ and $K_i = h_2[f_{En}(p_s^i)]$, and calculate an attention value as \cite{vaswani2017attention,2018BERT}
\begin{equation}
\text{att}_i = \text{softmax}(\frac{Q_iK_i^T}{\sqrt{d}}),
\end{equation}
where $d$ is the feature dimension of $Q_i$ and $K_i$. To ensure that the semantic information of the simple prompt is not directly interfered with, we directly set $V_i = f_{En}(p_s^i)$ without any transformation.  In particular, to embed the powerful semantic understanding and reasoning capabilities of the LLM in $\text{att}_i$, we distill knowledge from LLM by the following loss function:
\begin{equation}
\ell_{\text{LLM}}(\phi) = \mathbf{KL}[\mathbf{W}_0 f_{\text{LLM}}(p_s^i)/\tau, Q_i/\tau],
\label{eq:lossllm}
\end{equation}
Here, $\tau$ is the temperature, which is typically set to 1, and $\mathbf{KL}$ is the KL divergence. $\mathbf{W}_0$ is a randomly initialized matrix using Kaiming initialization and is unlearnable, which ensures that semantic information of LLM is reserved as much as possible while aligning the dimensions between $f_{\text{LLM}}(p_s^i)$ and $Q$. Moreover, we obtain the calibrated semantic information as $V_i^\prime = V_i \otimes \text{att}_i$.  Finally, the output of the Adapter is transformed by the learnable transformation $g(\cdot;\phi_1)$ to obtain the output $c_{\text{LLM}}$ with LLM semantic capabilities as prior works~\cite{He2015DeepRL,Huang2019DIANetDA,Zhong2023SPEMSP}:
\begin{equation}
 g\left\{ g_{\text{Ada}}(f_{En}(p_s^i);\phi_2);\phi_1 \right\} =  g\left\{V_i^\prime + V_i + h_1[V_i^\prime + V_i]; \phi_1 \right\},
\end{equation}
and the semantic information input to the predictor is as follows:
\begin{equation}
c_{\text{LLM}}^\prime = \eta  \cdot c_{\text{LLM}} + (1-\eta) \cdot  f_{En}(p_s^i).
    \label{eq:predinput}
\end{equation}
where $\eta$ is a constant.
\subsubsection{Performance Maintenance of DMs During Fine-tuning}
To maintain the performance of the diffusion model during fine-tuning, we add varying levels of noise to the image $I_i$ by Eq.(\ref{eq:xt}), and feed the semantic information feature $c_{\text{LLM}}^\prime$ obtained from Eq.(\ref{eq:predinput}) to the predictor, guided by the simple prompt $p_s^i$. To ensure that the pre-trained diffusion model maintains sufficient denoising ability for new images $I_i$ during fine-tuning, we minimize the following loss function:
\begin{equation}
\ell_\text{simple}^t(\phi) = \mathbb{E} \| \epsilon - \hat{\epsilon} (\alpha_t I_i+\sigma_t\epsilon, t, c_{\text{LLM}}^\prime) \|_2^2,
\label{eq:mianloss}
\end{equation}
Furthermore, to ensure stable training of the added adapter and reduce its adverse impact on the pre-trained diffusion model during the early stage of training, we follow the setting of previous works~\cite{hu2021lora,Zhang2023AddingCC} by initializing all elements of the matrices in parameter $\phi_1$ to 0.

\subsubsection{Aligning the Representation between Complex Prompts and Simple Prompts}
From the description in Section \ref{sec:datasetsurd}, we know that image $I_i$ is a semantically correct and high-quality image generated by $p_c^i$. In order to generate images of sufficient similarity and quality as $I_i$ by a simple prompt. we need to align the semantic representation of feature between $c_{\text{LLM}}^\prime$ and $f_{En}(p_c^i)$. Specifically, we consider minimizing the following loss function:
\begin{equation}
\ell_{\text{CP}}(\phi) = \mathbf{KL}(c_{\text{LLM}}^\prime/\tau, f_{En}(p_c^i)/\tau),
\label{eq:losscp}
\end{equation}
where $\tau$ is set as in Eq.(\ref{eq:lossllm}) and $\mathbf{KL}$ denotes the KL divergence~\cite{kullback1951information}.

In summary, the final loss function for SUR-adapter training is as follows:
\begin{equation}
\ell_{\text{total}}(\phi) = \lambda_1\cdot \ell_{\text{LLM}}(\phi) + \lambda_2\cdot \ell_{\text{CP}}(\phi) + \ell_\text{simple}^t(\phi),
\label{eq:totalloss}
\end{equation}
where $\lambda_i \leq 1$, $i=1,2$ are loss coefficients. We present the training process of SUR-adapter in Algorithm \ref{alg:ean}. After training, the fine-tuned diffusion model can generate images using the same sampling method as before.

\begin{table*}[htbp]
  \centering
  \caption{Evaluation results of the diverse pre-trained models and controlled methods described in Section \ref{sec:imp} in terms of various quality metrics. We calculate the T-test for the means of two independent samples of scores, and if the resulting P-value is greater than 0.05, it implies that there is no significant difference between the NR scores of the baselines and SUR-adapter, indicating that their generation quality is comparable. }
  \vspace{-0.3cm}
  \resizebox*{0.9\linewidth}{!}{
    \begin{tabular}{llcccccccc}
    \toprule
    \multicolumn{1}{l}{\multirow{2}[4]{*}{\makecell{Pre-trained\\Model}}} & \multirow{2}[4]{*}{Controlled Method} & \multicolumn{2}{c}{BRISQUE} & \multicolumn{2}{c}{CLIP-IQA} & \multicolumn{2}{c}{MUSIQ} & \multicolumn{2}{c}{User Preference (\%)} \\
\cmidrule{3-10}          &       & Baseline & Ours (P > 0.05?) & Baseline & Ours (P > 0.05?) & Baseline & Ours (P > 0.05?) & Baseline & Ours \\
    \midrule
    \multirow{6}[2]{*}{\makecell{DM (1.5),\\LLM (13B)}} & -     & 13.85  & 14.78 (\checkmark) & 0.686 & 0.688 (\checkmark) & 67.38  & 67.04 (\checkmark)  & 48.31 & 51.69 \\
          & ControlNet (canny) & 22.68  & 25.15 ($\times$)  & 0.673 & 0.668 (\checkmark) & 67.41  & 67.14 (\checkmark)  & 49.81 & 50.19 \\
          & ControlNet (seg) & 39.86  & 42.12 (\checkmark)  & 0.662 & 0.668 (\checkmark) & 64.12  & 65.71 ($\times$)  & 53.56 & 46.44 \\
          & Prompt Weighting & 13.29  & 13.74 (\checkmark)  & 0.681 & 0.691 (\checkmark) & 66.97  & 67.02 (\checkmark)  & 47.94 & 52.06 \\
          & MultiDiffusion & 10.84  & 11.83 (\checkmark)  & 0.696 & 0.691 (\checkmark) & 66.60  & 67.95 (\checkmark)  & 52.06 & 47.94  \\
          & Self-attention Guidance & 15.08  & 17.06 (\checkmark)  & 0.694 & 0.706 (\checkmark) & 67.51  & 68.97 (\checkmark)  & 48.31 & 51.69 \\
    \midrule
    \multirow{6}[2]{*}{\makecell{DM (cartoon),\\LLM (13B)}} & -     & 15.74  & 19.53 ($\times$)  & 0.699 & 0.707 (\checkmark) & 66.07  & 67.03 (\checkmark)  & 50.94 & 49.06 \\
          & ControlNet (canny) & 18.68  & 18.49 (\checkmark)  & 0.697 & 0.696 (\checkmark) & 67.98  & 67.95 (\checkmark)  & 50.56 & 49.44 \\
          & ControlNet (seg) & 35.84  & 31.96 ($\times$)  & 0.710 & 0.701 (\checkmark) & 67.51  & 67.68 (\checkmark)  & 51.69 & 48.31 \\
          & Prompt Weighting & 17.62  & 19.12 (\checkmark)  & 0.714 & 0.698 (\checkmark) & 67.38  & 66.46 (\checkmark)  & 51.31 & 48.69 \\
          & MultiDiffusion & 14.88  & 15.96 (\checkmark)  & 0.709 & 0.711 (\checkmark) & 68.05  & 67.26 (\checkmark)  & 47.94 & 52.06 \\
          & Self-attention Guidance & 20.44  & 20.98 (\checkmark)  & 0.705 & 0.706 (\checkmark) & 67.74  & 66.90 (\checkmark)  & 52.43 & 47.57  \\
    \bottomrule
    \end{tabular}%
    }
  \label{tab:quality}%
  \vspace{-0.3cm}
\end{table*}%

\begin{table*}[htbp]
  \centering
  \caption{The performance of diffusion models under various LLM settings. Bold and underline indicate the optimal and suboptimal performance, respectively.}
  \vspace{-0.3cm}
  \resizebox*{0.735\linewidth}{!}{
    \begin{tabular}{llccccccc}
    \toprule
    \makecell{Pre-trained\\Model} & \makecell{LLM Layer or\\Controlled Method} & CLIP Score & Action (\%) & Color (\%) & Counting (\%) & BRISQUE & CLIP-IQA & MUSIQ \\
    \midrule
    \multirow{5}[2]{*}{\makecell{DM (1.5),\\LLM (13B)}} & \multicolumn{1}{c}{1} & 0.414  & 68.00 & 82.00 & 32.67 & 13.89  & 0.688 & 67.04  \\
          & \multicolumn{1}{c}{10} & \underline{0.502}  & 74.00 & 84.67 & \underline{34.00} & 15.28  & 0.694 & 68.04 \\
          & \multicolumn{1}{c}{20} & 0.496  & \underline{78.00} & 81.33 & 30.00 & 15.77  & 0.690 & 67.27  \\
          & \multicolumn{1}{c}{30} & 0.482  & 72.67 & \textbf{90.00} & 31.33 & 17.85  & 0.691 & 67.25  \\
          & \multicolumn{1}{c}{40} & \textbf{0.517 } & \textbf{80.67} & \underline{87.33} & \textbf{36.67} & 14.78  & 0.684 & 67.47  \\
    \midrule
    \multirow{5}[2]{*}{\makecell{DM (cartoon),\\LLM (13B)}} 
          & \multicolumn{1}{c}{1} & 0.387  & 70.00 & 79.33 & 26.67 & 15.94  & 0.707 & 67.04 \\
          & \multicolumn{1}{c}{10} & 0.434  & 72.67 & 82.67 & \underline{34.67} & 17.31  & 0.703 & 66.02  \\
          & \multicolumn{1}{c}{20} & \underline{0.493}  & \underline{76.00} & 87.33 & 31.33 & 16.20  & 0.707 & 67.03  \\
          & \multicolumn{1}{c}{30} & \textbf{0.533 } & \textbf{78.00} & \textbf{91.33} & \textbf{38.00} & 17.58  & 0.707 & 66.50  \\
          & \multicolumn{1}{c}{40} & 0.490  & 68.67 & \underline{88.00} & \textbf{38.00} & 19.53 & 0.695 & 66.04  \\
    \midrule
    \multirow{6}[2]{*}{\makecell{DM (1.5),\\LLM (7B)}} & -     & 0.494  & 80.67 & 85.33 & 35.33 & 12.96  & 0.688 & 67.33  \\
          & ControlNet (canny) & 0.476  & 82.67 & 68.67 & 88.00 & 22.80  & 0.675 & 67.30  \\
          & ControlNet (seg) & 0.519  & 8.00  & 8.67  & 60.67 & 39.11  & 0.670 & 65.54  \\
          & Prompt Weighting & 0.601  & 84.00 & 83.33 & 53.33 & 14.53  & 0.688 & 67.09  \\
          & MultiDiffusion & 0.399  & 92.00 & 88.00 & 63.33 & 14.70  & 0.691 & 67.85  \\
          & Self-attention Guidance & 0.514  & 80.67 & 85.33 & 18.00 & 17.76  & 0.694 & 67.15  \\
    \midrule
    \multirow{6}[2]{*}{\makecell{DM (1.5),\\LLM (33B)}} & -     & 0.523  & 82.00 & 88.67 & 38.67 & 14.38  & 0.690 & 67.66  \\
          & ControlNet (canny) & 0.482  & 84.67 & 70.00 & 94.67 & 26.94  & 0.671 & 67.74  \\
          & ControlNet (seg) & 0.505  & 7.33  & 8.00  & 64.00 & 39.39  & 0.673 & 65.54  \\
          & Prompt Weighting & 0.530  & 84.67 & 92.67 & 58.67 & 13.96  & 0.702 & 67.38  \\
          & MultiDiffusion & 0.496  & 87.33 & 88.67 & 61.33 & 13.84  & 0.705 & 67.92  \\
          & Self-attention Guidance & 0.517  & 86.00 & 89.33 & 20.67 & 15.48  & 0.706 & 67.95  \\
    \bottomrule
    \end{tabular}%
    }
  \label{tab:llm}%
  \vspace{-0.3cm}
\end{table*}%

\section{Experiments}
\label{sec:exp}

\subsection{Implementation Details}
\label{sec:imp}

We utilize two pre-trained diffusion models (DMs) and three LLMs \cite{touvron2023llama} with different parameters. DM (1.5) \cite{Rombach2021HighResolutionIS} specialized in high-resolution image synthesis and DM (cartoon) \footnote{https://huggingface.co/nitrosocke/Ghibli-Diffusion} trained on modern anime feature film images. LLM ($s$) means the LLaMa model with the parameter size of $s$. In addition, we validate the universality of SUR-adapter with various controlled methods. ControlNet \cite{Zhang2023AddingCC} is an auxiliary network that introduces an additional condition. Our experiments include 2 canonical pre-trained ControlNets, namely edge detection with ControlNet (canny) and semantic segmentations with ControlNet (seg). Prompt weighting \footnote{https://github.com/damian0815/compel} is a straightforward technique that assigns higher attention weights to specific parts of the text input. MultiDiffusion \cite{bar2023multidiffusion} defines a novel generation process on top of a pre-trained diffusion model, which merges multiple diffusion generation methods. Self-attention Guidance \cite{hong2022improving} provides direction from predictions that are not reliant on high-frequency details to fully conditioned images. The high-frequency details are extracted from the UNet self-attention maps.

We use the SURD dataset to evaluate models by two types of metrics: semantic and quality. It is worth noting that all metrics are positively oriented. For \textbf{semantic evaluation}, we design three types of prompts \cite{Chen2022RethinkingDA,Chen2020CounterfactualSS,Agrawal2015VQAVQ,Liu2021SlakeAS}, namely Action, Color, and Counting, each with fifteen prompts. These prompts are used to evaluate the semantic capabilities of the baselines and SUR-adapter. Action, Color, and Counting are all percentage metrics that indicate the proportion of images that meet the different types of semantics. During testing, we generate ten images for each prompt. To further evaluate the semantic quality, we also use the CLIP Score \cite{radford2021learning}. We use CLIP to construct the binary classification problem for both the baselines and SUR-adapter and select the most appropriate images based on the prompts. After applying Softmax to avoid the effects of extreme values, we record the scores of the baselines and SUR-adapter, and use the mean value on the test set as the final CLIP score of the diffusion models. For \textbf{quality evaluation}, we use BRISQUE \cite{mittal2012no}, CLIP-IQA \cite{wang2022exploring}, MUSIQ \cite{ke2021musiq}, and user preference study. The user preference study consists of single-choice questions where users choose the image with the best quality from a pair of images generated by the baselines and SUR-adapter. We collected 89 valid questionnaires from the user preference study. In the appendix, we provide detailed training recipes.

\vspace{-0.3cm}
\subsection{Experiment Analysis}
Table \ref{tab:semanteme} shows the \textbf{semantic capabilities} of both baselines and SUR-adapter. Notably, the results demonstrate that SUR-adapter can effectively enhance the SUR performance of the baselines in most cases. Furthermore, we can draw the following conclusions: (a) the use of Softmax to obtain a relative score in CLIP can render the CLIP Score unreliable, particularly when both the baselines and SUR-adapter yield equally poor results. For instance, ControlNet (seg) attains a relatively high score despite its subpar generation effects on Action and Color. (b) ControlNet performs well in Counting scores since it utilizes image outlines with the correct amount of information as a reference. (c) Inaccurate image segmentation can cause diffusion models with ControlNet (seg) to disregard semantic information and generate entirely blurry images, thus resulting in unsatisfactory generation effects on Action and Color. Nonetheless, the negative impact of ControlNet (seg) can be alleviated by SUR-adapter. (d) The SUR capability of pre-trained diffusion models can be improved by employing Prompt Weighting and MultiDiffusion, with further enhancement achievable through the use of SUR-adapter.

As shown in Fig. \ref{fig:jg}, while the extra added adapter helps enhance the semantic understanding and reasoning abilities of diffusion models, adding additional parameters does not guarantee the preservation of the original image generation quality of the pre-trained diffusion models, as the adapter and pre-trained models are not trained simultaneously. However, our proposed SURD can be used to mitigate this issue by supplying high-quality images.  In Table \ref{tab:quality}, we demonstrate through multiple image quality metrics with T-tests and user preference study that SUR-adapter can maintain image generation quality, meaning there is no significant difference between the image quality of SUR-adapter and the original pre-trained diffusion model (P-value $\geq$ 0.05). Moreover, since these high-quality images of SURD also come from diffusion models, they do not lead to the generation of images of higher quality than those generated by the pre-trained diffusion models in our method.

\section{Ablation study}
\label{sec:ablation}

\noindent\textbf{The Analysis of LLMs. }As introduced in Section \ref{sec:data_ana}, LLM (13B) has 40 layers. The performance of LLM vectors with different layers is shown in the first two rows of Table \ref{tab:llm}. We find that in most cases, LLM vectors corresponding to the later layers are better. This suggests that the high-level semantic features in the deeper layers are more conducive to semantic distillation. Additionally, we show in the last two rows of Table \ref{tab:llm} the performance of LLMs with different parameter sizes. Combining the analysis of Table \ref{tab:llm}, \ref{tab:semanteme}, and \ref{tab:quality}, we find that there is no significant difference in diffusion model performance among LLMs with different parameter sizes. Although existing work suggests that models with larger parameter sizes have stronger SUR abilities, existing SUR-adapter may only be able to transfer limited semantic knowledge from LLMs.

\noindent\textbf{The Knowledge Distillation of SUR-adapter. }As shown in Table \ref{tab:distillation}, we conduct ablation studies on the knowledge distillation of LLM represented by the green line and complex prompts represented by the purple line in Fig. \ref{fig:jg}. Distilling the knowledge of LLM or complex prompts alone improves the SUR capability of SUR-adapter, and the effect of knowledge distillation based on LLM is stronger than that based on complex prompts. Furthermore, distilling the knowledge of both can further enhance the performance of SUR-adapter.

\begin{table}[htbp]
  \vspace{-0.1cm}
  \centering
  \caption{Ablation study on the knowledge distillation of SUR-adapter. }
  \vspace{-0.3cm}
  \resizebox*{0.85\linewidth}{!}{
    \begin{tabular}{cccccc}
    \toprule
    LLM   & \makecell{Complex\\Prompts} & BRISQUE & Action (\%) & Color (\%) & Number (\%) \\
    \midrule
          &       & 13.85 & 75.33 & 81.33 & 14.67 \\
    \checkmark     &       & 13.97 & 78.67 & 84.00 & 34.67 \\
          & \checkmark     & 12.31 & 74.00 & 86.67 & 32.00 \\
    \checkmark     & \checkmark     & 14.78 & \textbf{80.67} & \textbf{87.33} & \textbf{36.67} \\
    \bottomrule
    \end{tabular}%
    }
  \label{tab:distillation}%
  \vspace{-0.3cm}
\end{table}%

\section{Limitations}
As shown in Table \ref{tab:semanteme}, SUR-adapter has limited capacity to improve diffusion models and cannot completely address the SUR issue. For instance, after improvement, the Counting of DM (1.5), LLM (13B) is only increased by 36.67\%. However, addressing the deficiency of SUR may require a large-scale multimodal dataset to optimize the text encoder of diffusion models, which is a costly and challenging task. Moreover, as highlighted in Section \ref{sec:ablation}, there is no significant difference in performance among LLMs of different parameter sizes after distilling, indicating that SUR-adapter can only transfer limited semantic knowledge from LLMs due to factors such as parameter limitations. Hence, further enhancements are necessary for SUR-adapter to more effectively distill semantic information from LLMs.

\section{Conclusion}

In this paper, we uncover the limitations of existing pre-trained diffusion models in terms of their ability to comprehend semantics and engage in commonsense reasoning when presented with simple narrative prompts as inputs, leading to suboptimal image generation. To mitigate this issue, we introduce a new dataset called SURD, which comprises over 57,000 semantically corrected image-text pairs, and the SUR-adapter module that can distill semantic understanding and reasoning knowledge from complex keyword-based prompts and large language models. Extensive experiments and rigorous evaluations conducted on SURD demonstrate that SUR-adapter can enhance the semantic understanding of diffusion models without compromising image generation quality.

\begin{acks}
This work was supported in part by National Natural Science Foundation of China (NSFC) under Grant No.62206314 and Grant No.U1711264, GuangDong Basic and Applied Basic Research Foundation under Grant No.2022A1515011835, China Postdoctoral Science Foundation funded project under Grant No.2021M703687.
\end{acks}

\clearpage
\bibliographystyle{ACM-Reference-Format}
\balance
\bibliography{sample-base}


\begin{thebibliography}{58}

%%% the Web address.

\ifx \showCODEN    \undefined \def \showCODEN     #1{\unskip}     \fi
\ifx \showDOI      \undefined \def \showDOI       #1{#1}\fi
\ifx \showISBNx    \undefined \def \showISBNx     #1{\unskip}     \fi
\ifx \showISBNxiii \undefined \def \showISBNxiii  #1{\unskip}     \fi
\ifx \showISSN     \undefined \def \showISSN      #1{\unskip}     \fi
\ifx \showLCCN     \undefined \def \showLCCN      #1{\unskip}     \fi
\ifx \shownote     \undefined \def \shownote      #1{#1}          \fi
\ifx \showarticletitle \undefined \def \showarticletitle #1{#1}   \fi
\ifx \showURL      \undefined \def \showURL       {\relax}        \fi
% invisible to TeX
\providecommand\bibfield[2]{#2}
\providecommand\bibinfo[2]{#2}
\providecommand\natexlab[1]{#1}
\providecommand\showeprint[2][]{arXiv:#2}

\bibitem[Agrawal et~al\mbox{.}(2015)]%
        {Agrawal2015VQAVQ}
\bibfield{author}{\bibinfo{person}{Aishwarya Agrawal}, \bibinfo{person}{Jiasen
  Lu}, \bibinfo{person}{Stanislaw Antol}, \bibinfo{person}{Margaret Mitchell},
  \bibinfo{person}{C.~Lawrence Zitnick}, \bibinfo{person}{Devi Parikh}, {and}
  \bibinfo{person}{Dhruv Batra}.} \bibinfo{year}{2015}\natexlab{}.
\newblock \showarticletitle{VQA: Visual Question Answering}.
\newblock \bibinfo{journal}{\emph{International Journal of Computer Vision}}
  \bibinfo{volume}{123} (\bibinfo{year}{2015}), \bibinfo{pages}{4--31}.
\newblock


\bibitem[Bao et~al\mbox{.}(2023)]%
        {Bao2023OneTF}
\bibfield{author}{\bibinfo{person}{Fan Bao}, \bibinfo{person}{Shen Nie},
  \bibinfo{person}{Kaiwen Xue}, \bibinfo{person}{Chongxuan Li},
  \bibinfo{person}{Shiliang Pu}, \bibinfo{person}{Yaole Wang},
  \bibinfo{person}{Gang Yue}, \bibinfo{person}{Yue Cao}, \bibinfo{person}{Hang
  Su}, {and} \bibinfo{person}{Jun Zhu}.} \bibinfo{year}{2023}\natexlab{}.
\newblock \showarticletitle{One Transformer Fits All Distributions in
  Multi-Modal Diffusion at Scale}.
\newblock \bibinfo{journal}{\emph{ArXiv}}  \bibinfo{volume}{abs/2303.06555}
  (\bibinfo{year}{2023}).
\newblock


\bibitem[Bar-Tal et~al\mbox{.}(2023)]%
        {bar2023multidiffusion}
\bibfield{author}{\bibinfo{person}{Omer Bar-Tal}, \bibinfo{person}{Lior Yariv},
  \bibinfo{person}{Yaron Lipman}, {and} \bibinfo{person}{Tali Dekel}.}
  \bibinfo{year}{2023}\natexlab{}.
\newblock \showarticletitle{MultiDiffusion: Fusing Diffusion Paths for
  Controlled Image Generation}.
\newblock \bibinfo{journal}{\emph{arXiv preprint arXiv:2302.08113}}
  \bibinfo{volume}{2} (\bibinfo{year}{2023}).
\newblock


\bibitem[Baranchuk et~al\mbox{.}(2021)]%
        {Baranchuk2021LabelEfficientSS}
\bibfield{author}{\bibinfo{person}{Dmitry Baranchuk}, \bibinfo{person}{Ivan
  Rubachev}, \bibinfo{person}{Andrey Voynov}, \bibinfo{person}{Valentin
  Khrulkov}, {and} \bibinfo{person}{Artem Babenko}.}
  \bibinfo{year}{2021}\natexlab{}.
\newblock \showarticletitle{Label-Efficient Semantic Segmentation with
  Diffusion Models}.
\newblock \bibinfo{journal}{\emph{ArXiv}}  \bibinfo{volume}{abs/2112.03126}
  (\bibinfo{year}{2021}).
\newblock


\bibitem[Batzolis et~al\mbox{.}(2021)]%
        {Batzolis2021ConditionalIG}
\bibfield{author}{\bibinfo{person}{Georgios Batzolis}, \bibinfo{person}{Jan
  Stanczuk}, \bibinfo{person}{Carola-Bibiane Schonlieb}, {and}
  \bibinfo{person}{Christian Etmann}.} \bibinfo{year}{2021}\natexlab{}.
\newblock \showarticletitle{Conditional Image Generation with Score-Based
  Diffusion Models}.
\newblock


\bibitem[Brown et~al\mbox{.}(2020)]%
        {Brown2020LanguageMA}
\bibfield{author}{\bibinfo{person}{Tom~B. Brown}, \bibinfo{person}{Benjamin
  Mann}, \bibinfo{person}{Nick Ryder}, \bibinfo{person}{Melanie Subbiah},
  \bibinfo{person}{Jared Kaplan}, \bibinfo{person}{Prafulla Dhariwal},
  \bibinfo{person}{Arvind Neelakantan}, \bibinfo{person}{Pranav Shyam},
  \bibinfo{person}{Girish Sastry}, \bibinfo{person}{Amanda Askell},
  \bibinfo{person}{Sandhini Agarwal}, \bibinfo{person}{Ariel Herbert-Voss},
  \bibinfo{person}{Gretchen Krueger}, \bibinfo{person}{T.~J. Henighan},
  \bibinfo{person}{Rewon Child}, \bibinfo{person}{Aditya Ramesh},
  \bibinfo{person}{Daniel~M. Ziegler}, \bibinfo{person}{Jeff Wu},
  \bibinfo{person}{Clemens Winter}, \bibinfo{person}{Christopher Hesse},
  \bibinfo{person}{Mark Chen}, \bibinfo{person}{Eric Sigler},
  \bibinfo{person}{Mateusz Litwin}, \bibinfo{person}{Scott Gray},
  \bibinfo{person}{Benjamin Chess}, \bibinfo{person}{Jack Clark},
  \bibinfo{person}{Christopher Berner}, \bibinfo{person}{Sam McCandlish},
  \bibinfo{person}{Alec Radford}, \bibinfo{person}{Ilya Sutskever}, {and}
  \bibinfo{person}{Dario Amodei}.} \bibinfo{year}{2020}\natexlab{}.
\newblock \showarticletitle{Language Models are Few-Shot Learners}.
\newblock \bibinfo{journal}{\emph{ArXiv}}  \bibinfo{volume}{abs/2005.14165}
  (\bibinfo{year}{2020}).
\newblock


\bibitem[Chen et~al\mbox{.}(2020)]%
        {Chen2020CounterfactualSS}
\bibfield{author}{\bibinfo{person}{Long Chen}, \bibinfo{person}{Xin Yan},
  \bibinfo{person}{Jun Xiao}, \bibinfo{person}{Hanwang Zhang},
  \bibinfo{person}{Shiliang Pu}, {and} \bibinfo{person}{Yueting Zhuang}.}
  \bibinfo{year}{2020}\natexlab{}.
\newblock \showarticletitle{Counterfactual Samples Synthesizing for Robust
  Visual Question Answering}.
\newblock \bibinfo{journal}{\emph{2020 IEEE/CVF Conference on Computer Vision
  and Pattern Recognition (CVPR)}} (\bibinfo{year}{2020}),
  \bibinfo{pages}{10797--10806}.
\newblock


\bibitem[Chen et~al\mbox{.}(2022)]%
        {Chen2022RethinkingDA}
\bibfield{author}{\bibinfo{person}{Long Chen}, \bibinfo{person}{Yuhang Zheng},
  {and} \bibinfo{person}{Jun Xiao}.} \bibinfo{year}{2022}\natexlab{}.
\newblock \showarticletitle{Rethinking Data Augmentation for Robust Visual
  Question Answering}.
\newblock \bibinfo{journal}{\emph{ArXiv}}  \bibinfo{volume}{abs/2207.08739}
  (\bibinfo{year}{2022}).
\newblock


\bibitem[Chowdhery et~al\mbox{.}(2022)]%
        {Chowdhery2022PaLMSL}
\bibfield{author}{\bibinfo{person}{Aakanksha Chowdhery},
  \bibinfo{person}{Sharan Narang}, \bibinfo{person}{Jacob Devlin},
  \bibinfo{person}{Maarten Bosma}, \bibinfo{person}{Gaurav Mishra},
  \bibinfo{person}{Adam Roberts}, \bibinfo{person}{Paul Barham},
  \bibinfo{person}{Hyung~Won Chung}, \bibinfo{person}{Charles Sutton},
  \bibinfo{person}{Sebastian Gehrmann}, \bibinfo{person}{Parker Schuh},
  \bibinfo{person}{Kensen Shi}, \bibinfo{person}{Sasha Tsvyashchenko},
  \bibinfo{person}{Joshua Maynez}, \bibinfo{person}{Abhishek Rao},
  \bibinfo{person}{Parker Barnes}, \bibinfo{person}{Yi Tay},
  \bibinfo{person}{Noam~M. Shazeer}, \bibinfo{person}{Vinodkumar Prabhakaran},
  \bibinfo{person}{Emily Reif}, \bibinfo{person}{Nan Du},
  \bibinfo{person}{Benton~C. Hutchinson}, \bibinfo{person}{Reiner Pope},
  \bibinfo{person}{James Bradbury}, \bibinfo{person}{Jacob Austin},
  \bibinfo{person}{Michael Isard}, \bibinfo{person}{Guy Gur-Ari},
  \bibinfo{person}{Pengcheng Yin}, \bibinfo{person}{Toju Duke},
  \bibinfo{person}{Anselm Levskaya}, \bibinfo{person}{Sanjay Ghemawat},
  \bibinfo{person}{Sunipa Dev}, \bibinfo{person}{Henryk Michalewski},
  \bibinfo{person}{Xavier Garc{\'i}a}, \bibinfo{person}{Vedant Misra},
  \bibinfo{person}{Kevin Robinson}, \bibinfo{person}{Liam Fedus},
  \bibinfo{person}{Denny Zhou}, \bibinfo{person}{Daphne Ippolito},
  \bibinfo{person}{David Luan}, \bibinfo{person}{Hyeontaek Lim},
  \bibinfo{person}{Barret Zoph}, \bibinfo{person}{Alexander Spiridonov},
  \bibinfo{person}{Ryan Sepassi}, \bibinfo{person}{David Dohan},
  \bibinfo{person}{Shivani Agrawal}, \bibinfo{person}{Mark Omernick},
  \bibinfo{person}{Andrew~M. Dai},
  \bibinfo{person}{Thanumalayan~Sankaranarayana Pillai}, \bibinfo{person}{Marie
  Pellat}, \bibinfo{person}{Aitor Lewkowycz}, \bibinfo{person}{Erica Moreira},
  \bibinfo{person}{Rewon Child}, \bibinfo{person}{Oleksandr Polozov},
  \bibinfo{person}{Katherine Lee}, \bibinfo{person}{Zongwei Zhou},
  \bibinfo{person}{Xuezhi Wang}, \bibinfo{person}{Brennan Saeta},
  \bibinfo{person}{Mark D{\'i}az}, \bibinfo{person}{Orhan Firat},
  \bibinfo{person}{Michele Catasta}, \bibinfo{person}{Jason Wei},
  \bibinfo{person}{Kathleen~S. Meier-Hellstern}, \bibinfo{person}{Douglas Eck},
  \bibinfo{person}{Jeff Dean}, \bibinfo{person}{Slav Petrov}, {and}
  \bibinfo{person}{Noah Fiedel}.} \bibinfo{year}{2022}\natexlab{}.
\newblock \showarticletitle{PaLM: Scaling Language Modeling with Pathways}.
\newblock \bibinfo{journal}{\emph{ArXiv}}  \bibinfo{volume}{abs/2204.02311}
  (\bibinfo{year}{2022}).
\newblock


\bibitem[Devlin et~al\mbox{.}(2018)]%
        {2018BERT}
\bibfield{author}{\bibinfo{person}{J. Devlin}, \bibinfo{person}{M.~W. Chang},
  \bibinfo{person}{K. Lee}, {and} \bibinfo{person}{K. Toutanova}.}
  \bibinfo{year}{2018}\natexlab{}.
\newblock \showarticletitle{BERT: Pre-training of Deep Bidirectional
  Transformers for Language Understanding}.
\newblock  (\bibinfo{year}{2018}).
\newblock


\bibitem[Fan et~al\mbox{.}(2022)]%
        {Fan2022FridoFP}
\bibfield{author}{\bibinfo{person}{Wanshu Fan}, \bibinfo{person}{Yen-Chun
  Chen}, \bibinfo{person}{Dongdong Chen}, \bibinfo{person}{Yu Cheng},
  \bibinfo{person}{Lu Yuan}, {and} \bibinfo{person}{Yu-Chiang~Frank Wang}.}
  \bibinfo{year}{2022}\natexlab{}.
\newblock \showarticletitle{Frido: Feature Pyramid Diffusion for Complex Scene
  Image Synthesis}.
\newblock \bibinfo{journal}{\emph{ArXiv}}  \bibinfo{volume}{abs/2208.13753}
  (\bibinfo{year}{2022}).
\newblock


\bibitem[Graikos et~al\mbox{.}(2022)]%
        {Graikos2022DiffusionMA}
\bibfield{author}{\bibinfo{person}{Alexandros Graikos},
  \bibinfo{person}{Nikolay Malkin}, \bibinfo{person}{Nebojsa Jojic}, {and}
  \bibinfo{person}{Dimitris Samaras}.} \bibinfo{year}{2022}\natexlab{}.
\newblock \showarticletitle{Diffusion models as plug-and-play priors}.
\newblock \bibinfo{journal}{\emph{ArXiv}}  \bibinfo{volume}{abs/2206.09012}
  (\bibinfo{year}{2022}).
\newblock


\bibitem[Guo et~al\mbox{.}(2022)]%
        {guo2022images}
\bibfield{author}{\bibinfo{person}{Jiaxian Guo}, \bibinfo{person}{Junnan Li},
  \bibinfo{person}{Dongxu Li}, \bibinfo{person}{Anthony Meng~Huat Tiong},
  \bibinfo{person}{Boyang Li}, \bibinfo{person}{Dacheng Tao}, {and}
  \bibinfo{person}{Steven~CH Hoi}.} \bibinfo{year}{2022}\natexlab{}.
\newblock \showarticletitle{From Images to Textual Prompts: Zero-shot VQA with
  Frozen Large Language Models}.
\newblock \bibinfo{journal}{\emph{arXiv preprint arXiv:2212.10846}}
  (\bibinfo{year}{2022}).
\newblock


\bibitem[He et~al\mbox{.}(2015)]%
        {He2015DeepRL}
\bibfield{author}{\bibinfo{person}{Kaiming He}, \bibinfo{person}{X. Zhang},
  \bibinfo{person}{Shaoqing Ren}, {and} \bibinfo{person}{Jian Sun}.}
  \bibinfo{year}{2015}\natexlab{}.
\newblock \showarticletitle{Deep Residual Learning for Image Recognition}.
\newblock \bibinfo{journal}{\emph{2016 IEEE Conference on Computer Vision and
  Pattern Recognition (CVPR)}} (\bibinfo{year}{2015}),
  \bibinfo{pages}{770--778}.
\newblock


\bibitem[Hestness et~al\mbox{.}(2017)]%
        {Hestness2017DeepLS}
\bibfield{author}{\bibinfo{person}{Joel Hestness}, \bibinfo{person}{Sharan
  Narang}, \bibinfo{person}{Newsha Ardalani},
  \bibinfo{person}{Gregory~Frederick Diamos}, \bibinfo{person}{Heewoo Jun},
  \bibinfo{person}{Hassan Kianinejad}, \bibinfo{person}{Md. Mostofa~Ali
  Patwary}, \bibinfo{person}{Yang Yang}, {and} \bibinfo{person}{Yanqi Zhou}.}
  \bibinfo{year}{2017}\natexlab{}.
\newblock \showarticletitle{Deep Learning Scaling is Predictable, Empirically}.
\newblock \bibinfo{journal}{\emph{ArXiv}}  \bibinfo{volume}{abs/1712.00409}
  (\bibinfo{year}{2017}).
\newblock


\bibitem[Ho et~al\mbox{.}(2020)]%
        {Ho2020DenoisingDP}
\bibfield{author}{\bibinfo{person}{Jonathan Ho}, \bibinfo{person}{Ajay Jain},
  {and} \bibinfo{person}{P. Abbeel}.} \bibinfo{year}{2020}\natexlab{}.
\newblock \showarticletitle{Denoising Diffusion Probabilistic Models}.
\newblock \bibinfo{journal}{\emph{ArXiv}}  \bibinfo{volume}{abs/2006.11239}
  (\bibinfo{year}{2020}).
\newblock


\bibitem[Hoffmann et~al\mbox{.}(2022)]%
        {Hoffmann2022TrainingCL}
\bibfield{author}{\bibinfo{person}{Jordan Hoffmann}, \bibinfo{person}{Sebastian
  Borgeaud}, \bibinfo{person}{Arthur Mensch}, \bibinfo{person}{Elena
  Buchatskaya}, \bibinfo{person}{Trevor Cai}, \bibinfo{person}{Eliza
  Rutherford}, \bibinfo{person}{Diego de Las~Casas}, \bibinfo{person}{Lisa~Anne
  Hendricks}, \bibinfo{person}{Johannes Welbl}, \bibinfo{person}{Aidan Clark},
  \bibinfo{person}{Tom Hennigan}, \bibinfo{person}{Eric Noland},
  \bibinfo{person}{Katie Millican}, \bibinfo{person}{George van~den Driessche},
  \bibinfo{person}{Bogdan Damoc}, \bibinfo{person}{Aurelia Guy},
  \bibinfo{person}{Simon Osindero}, \bibinfo{person}{Karen Simonyan},
  \bibinfo{person}{Erich Elsen}, \bibinfo{person}{Jack~W. Rae},
  \bibinfo{person}{Oriol Vinyals}, {and} \bibinfo{person}{L. Sifre}.}
  \bibinfo{year}{2022}\natexlab{}.
\newblock \showarticletitle{Training Compute-Optimal Large Language Models}.
\newblock \bibinfo{journal}{\emph{ArXiv}}  \bibinfo{volume}{abs/2203.15556}
  (\bibinfo{year}{2022}).
\newblock


\bibitem[Hong et~al\mbox{.}(2022)]%
        {hong2022improving}
\bibfield{author}{\bibinfo{person}{Susung Hong}, \bibinfo{person}{Gyuseong
  Lee}, \bibinfo{person}{Wooseok Jang}, {and} \bibinfo{person}{Seungryong
  Kim}.} \bibinfo{year}{2022}\natexlab{}.
\newblock \showarticletitle{Improving Sample Quality of Diffusion Models Using
  Self-Attention Guidance}.
\newblock \bibinfo{journal}{\emph{arXiv preprint arXiv:2210.00939}}
  (\bibinfo{year}{2022}).
\newblock


\bibitem[Hu et~al\mbox{.}(2021)]%
        {hu2021lora}
\bibfield{author}{\bibinfo{person}{Edward~J Hu}, \bibinfo{person}{Yelong Shen},
  \bibinfo{person}{Phillip Wallis}, \bibinfo{person}{Zeyuan Allen-Zhu},
  \bibinfo{person}{Yuanzhi Li}, \bibinfo{person}{Shean Wang},
  \bibinfo{person}{Lu Wang}, {and} \bibinfo{person}{Weizhu Chen}.}
  \bibinfo{year}{2021}\natexlab{}.
\newblock \showarticletitle{Lora: Low-rank adaptation of large language
  models}.
\newblock \bibinfo{journal}{\emph{arXiv preprint arXiv:2106.09685}}
  (\bibinfo{year}{2021}).
\newblock


\bibitem[Huang et~al\mbox{.}(2019)]%
        {Huang2019DIANetDA}
\bibfield{author}{\bibinfo{person}{Zhongzhan Huang}, \bibinfo{person}{Senwei
  Liang}, \bibinfo{person}{Mingfu Liang}, {and} \bibinfo{person}{Haizhao
  Yang}.} \bibinfo{year}{2019}\natexlab{}.
\newblock \showarticletitle{DIANet: Dense-and-Implicit Attention Network}. In
  \bibinfo{booktitle}{\emph{AAAI Conference on Artificial Intelligence}}.
\newblock


\bibitem[Jozefowicz et~al\mbox{.}(2016)]%
        {2016Exploring}
\bibfield{author}{\bibinfo{person}{R. Jozefowicz}, \bibinfo{person}{O.
  Vinyals}, \bibinfo{person}{M. Schuster}, \bibinfo{person}{N. Shazeer}, {and}
  \bibinfo{person}{Y. Wu}.} \bibinfo{year}{2016}\natexlab{}.
\newblock \bibinfo{title}{Exploring the Limits of Language Modeling}.
\newblock
\newblock


\bibitem[Kaplan et~al\mbox{.}(2020)]%
        {Kaplan2020ScalingLF}
\bibfield{author}{\bibinfo{person}{Jared Kaplan}, \bibinfo{person}{Sam
  McCandlish}, \bibinfo{person}{T.~J. Henighan}, \bibinfo{person}{Tom~B.
  Brown}, \bibinfo{person}{Benjamin Chess}, \bibinfo{person}{Rewon Child},
  \bibinfo{person}{Scott Gray}, \bibinfo{person}{Alec Radford},
  \bibinfo{person}{Jeff Wu}, {and} \bibinfo{person}{Dario Amodei}.}
  \bibinfo{year}{2020}\natexlab{}.
\newblock \showarticletitle{Scaling Laws for Neural Language Models}.
\newblock \bibinfo{journal}{\emph{ArXiv}}  \bibinfo{volume}{abs/2001.08361}
  (\bibinfo{year}{2020}).
\newblock


\bibitem[Kawar et~al\mbox{.}(2022)]%
        {Kawar2022ImagicTR}
\bibfield{author}{\bibinfo{person}{Bahjat Kawar}, \bibinfo{person}{Shiran
  Zada}, \bibinfo{person}{Oran Lang}, \bibinfo{person}{Omer Tov},
  \bibinfo{person}{Hui-Tang Chang}, \bibinfo{person}{Tali Dekel},
  \bibinfo{person}{Inbar Mosseri}, {and} \bibinfo{person}{Michal Irani}.}
  \bibinfo{year}{2022}\natexlab{}.
\newblock \showarticletitle{Imagic: Text-Based Real Image Editing with
  Diffusion Models}.
\newblock \bibinfo{journal}{\emph{ArXiv}}  \bibinfo{volume}{abs/2210.09276}
  (\bibinfo{year}{2022}).
\newblock


\bibitem[Ke et~al\mbox{.}(2021)]%
        {ke2021musiq}
\bibfield{author}{\bibinfo{person}{Junjie Ke}, \bibinfo{person}{Qifei Wang},
  \bibinfo{person}{Yilin Wang}, \bibinfo{person}{Peyman Milanfar}, {and}
  \bibinfo{person}{Feng Yang}.} \bibinfo{year}{2021}\natexlab{}.
\newblock \showarticletitle{Musiq: Multi-scale image quality transformer}. In
  \bibinfo{booktitle}{\emph{Proceedings of the IEEE/CVF International
  Conference on Computer Vision}}. \bibinfo{pages}{5148--5157}.
\newblock


\bibitem[Kim et~al\mbox{.}(2021)]%
        {Kim2021DiffusionCLIPTD}
\bibfield{author}{\bibinfo{person}{Gwanghyun Kim}, \bibinfo{person}{Taesung
  Kwon}, {and} \bibinfo{person}{Jong-Chul Ye}.}
  \bibinfo{year}{2021}\natexlab{}.
\newblock \showarticletitle{DiffusionCLIP: Text-Guided Diffusion Models for
  Robust Image Manipulation}.
\newblock \bibinfo{journal}{\emph{2022 IEEE/CVF Conference on Computer Vision
  and Pattern Recognition (CVPR)}} (\bibinfo{year}{2021}),
  \bibinfo{pages}{2416--2425}.
\newblock


\bibitem[Kullback and Leibler(1951)]%
        {kullback1951information}
\bibfield{author}{\bibinfo{person}{Solomon Kullback} {and}
  \bibinfo{person}{Richard~A Leibler}.} \bibinfo{year}{1951}\natexlab{}.
\newblock \showarticletitle{On information and sufficiency}.
\newblock \bibinfo{journal}{\emph{The annals of mathematical statistics}}
  \bibinfo{volume}{22}, \bibinfo{number}{1} (\bibinfo{year}{1951}),
  \bibinfo{pages}{79--86}.
\newblock


\bibitem[Li et~al\mbox{.}(2021)]%
        {Li2021SRDiffSI}
\bibfield{author}{\bibinfo{person}{Haoying Li}, \bibinfo{person}{Yifan Yang},
  \bibinfo{person}{Meng Chang}, \bibinfo{person}{Huajun Feng},
  \bibinfo{person}{Zhi hai Xu}, \bibinfo{person}{Qi Li}, {and}
  \bibinfo{person}{Yue ting Chen}.} \bibinfo{year}{2021}\natexlab{}.
\newblock \showarticletitle{SRDiff: Single Image Super-Resolution with
  Diffusion Probabilistic Models}.
\newblock \bibinfo{journal}{\emph{Neurocomputing}}  \bibinfo{volume}{479}
  (\bibinfo{year}{2021}), \bibinfo{pages}{47--59}.
\newblock


\bibitem[Li et~al\mbox{.}(2022)]%
        {li2022blip}
\bibfield{author}{\bibinfo{person}{Junnan Li}, \bibinfo{person}{Dongxu Li},
  \bibinfo{person}{Caiming Xiong}, {and} \bibinfo{person}{Steven Hoi}.}
  \bibinfo{year}{2022}\natexlab{}.
\newblock \showarticletitle{Blip: Bootstrapping language-image pre-training for
  unified vision-language understanding and generation}. In
  \bibinfo{booktitle}{\emph{International Conference on Machine Learning}}.
  PMLR, \bibinfo{pages}{12888--12900}.
\newblock


\bibitem[Lieber et~al\mbox{.}(2021)]%
        {lieber2021jurassic}
\bibfield{author}{\bibinfo{person}{Opher Lieber}, \bibinfo{person}{Or Sharir},
  \bibinfo{person}{Barak Lenz}, {and} \bibinfo{person}{Yoav Shoham}.}
  \bibinfo{year}{2021}\natexlab{}.
\newblock \showarticletitle{Jurassic-1: Technical details and evaluation}.
\newblock \bibinfo{journal}{\emph{White Paper. AI21 Labs}}  \bibinfo{volume}{1}
  (\bibinfo{year}{2021}).
\newblock


\bibitem[Lin et~al\mbox{.}(2014)]%
        {Lin2014MicrosoftCC}
\bibfield{author}{\bibinfo{person}{Tsung-Yi Lin}, \bibinfo{person}{Michael
  Maire}, \bibinfo{person}{Serge~J. Belongie}, \bibinfo{person}{James Hays},
  \bibinfo{person}{Pietro Perona}, \bibinfo{person}{Deva Ramanan},
  \bibinfo{person}{Piotr Doll{\'a}r}, {and} \bibinfo{person}{C.~Lawrence
  Zitnick}.} \bibinfo{year}{2014}\natexlab{}.
\newblock \showarticletitle{Microsoft COCO: Common Objects in Context}. In
  \bibinfo{booktitle}{\emph{European Conference on Computer Vision}}.
\newblock


\bibitem[Liu et~al\mbox{.}(2021)]%
        {Liu2021SlakeAS}
\bibfield{author}{\bibinfo{person}{Bo Liu}, \bibinfo{person}{Li-Ming Zhan},
  \bibinfo{person}{Li Xu}, \bibinfo{person}{Lin Ma}, \bibinfo{person}{Yan~Fang
  Yang}, {and} \bibinfo{person}{Xiao-Ming Wu}.}
  \bibinfo{year}{2021}\natexlab{}.
\newblock \showarticletitle{Slake: A Semantically-Labeled Knowledge-Enhanced
  Dataset For Medical Visual Question Answering}.
\newblock \bibinfo{journal}{\emph{2021 IEEE 18th International Symposium on
  Biomedical Imaging (ISBI)}} (\bibinfo{year}{2021}),
  \bibinfo{pages}{1650--1654}.
\newblock


\bibitem[Lugmayr et~al\mbox{.}(2022)]%
        {Lugmayr2022RePaintIU}
\bibfield{author}{\bibinfo{person}{Andreas Lugmayr}, \bibinfo{person}{Martin
  Danelljan}, \bibinfo{person}{Andr{\'e}s Romero}, \bibinfo{person}{Fisher Yu},
  \bibinfo{person}{Radu Timofte}, {and} \bibinfo{person}{Luc~Van Gool}.}
  \bibinfo{year}{2022}\natexlab{}.
\newblock \showarticletitle{RePaint: Inpainting using Denoising Diffusion
  Probabilistic Models}.
\newblock \bibinfo{journal}{\emph{2022 IEEE/CVF Conference on Computer Vision
  and Pattern Recognition (CVPR)}} (\bibinfo{year}{2022}),
  \bibinfo{pages}{11451--11461}.
\newblock


\bibitem[Mittal et~al\mbox{.}(2012)]%
        {mittal2012no}
\bibfield{author}{\bibinfo{person}{Anish Mittal},
  \bibinfo{person}{Anush~Krishna Moorthy}, {and} \bibinfo{person}{Alan~Conrad
  Bovik}.} \bibinfo{year}{2012}\natexlab{}.
\newblock \showarticletitle{No-reference image quality assessment in the
  spatial domain}.
\newblock \bibinfo{journal}{\emph{IEEE Transactions on image processing}}
  \bibinfo{volume}{21}, \bibinfo{number}{12} (\bibinfo{year}{2012}),
  \bibinfo{pages}{4695--4708}.
\newblock


\bibitem[Radford et~al\mbox{.}(2021)]%
        {radford2021learning}
\bibfield{author}{\bibinfo{person}{Alec Radford}, \bibinfo{person}{Jong~Wook
  Kim}, \bibinfo{person}{Chris Hallacy}, \bibinfo{person}{Aditya Ramesh},
  \bibinfo{person}{Gabriel Goh}, \bibinfo{person}{Sandhini Agarwal},
  \bibinfo{person}{Girish Sastry}, \bibinfo{person}{Amanda Askell},
  \bibinfo{person}{Pamela Mishkin}, \bibinfo{person}{Jack Clark},
  {et~al\mbox{.}}} \bibinfo{year}{2021}\natexlab{}.
\newblock \showarticletitle{Learning transferable visual models from natural
  language supervision}. In \bibinfo{booktitle}{\emph{International conference
  on machine learning}}. PMLR, \bibinfo{pages}{8748--8763}.
\newblock


\bibitem[Radford et~al\mbox{.}(2019)]%
        {Radford2019LanguageMA}
\bibfield{author}{\bibinfo{person}{Alec Radford}, \bibinfo{person}{Jeff Wu},
  \bibinfo{person}{Rewon Child}, \bibinfo{person}{David Luan},
  \bibinfo{person}{Dario Amodei}, {and} \bibinfo{person}{Ilya Sutskever}.}
  \bibinfo{year}{2019}\natexlab{}.
\newblock \showarticletitle{Language Models are Unsupervised Multitask
  Learners}.
\newblock


\bibitem[Rae et~al\mbox{.}(2021)]%
        {Rae2021ScalingLM}
\bibfield{author}{\bibinfo{person}{Jack~W. Rae}, \bibinfo{person}{Sebastian
  Borgeaud}, \bibinfo{person}{Trevor Cai}, \bibinfo{person}{Katie Millican},
  \bibinfo{person}{Jordan Hoffmann}, \bibinfo{person}{Francis Song},
  \bibinfo{person}{John Aslanides}, \bibinfo{person}{Sarah Henderson},
  \bibinfo{person}{Roman Ring}, \bibinfo{person}{Susannah Young},
  \bibinfo{person}{Eliza Rutherford}, \bibinfo{person}{Tom Hennigan},
  \bibinfo{person}{Jacob Menick}, \bibinfo{person}{Albin Cassirer},
  \bibinfo{person}{Richard Powell}, \bibinfo{person}{George van~den Driessche},
  \bibinfo{person}{Lisa~Anne Hendricks}, \bibinfo{person}{Maribeth Rauh},
  \bibinfo{person}{Po-Sen Huang}, \bibinfo{person}{Amelia Glaese},
  \bibinfo{person}{Johannes Welbl}, \bibinfo{person}{Sumanth Dathathri},
  \bibinfo{person}{Saffron Huang}, \bibinfo{person}{Jonathan Uesato},
  \bibinfo{person}{John F.~J. Mellor}, \bibinfo{person}{Irina Higgins},
  \bibinfo{person}{Antonia Creswell}, \bibinfo{person}{Nathan McAleese},
  \bibinfo{person}{Amy Wu}, \bibinfo{person}{Erich Elsen},
  \bibinfo{person}{Siddhant~M. Jayakumar}, \bibinfo{person}{Elena Buchatskaya},
  \bibinfo{person}{David Budden}, \bibinfo{person}{Esme Sutherland},
  \bibinfo{person}{Karen Simonyan}, \bibinfo{person}{Michela Paganini},
  \bibinfo{person}{L. Sifre}, \bibinfo{person}{Lena Martens},
  \bibinfo{person}{Xiang~Lorraine Li}, \bibinfo{person}{Adhiguna Kuncoro},
  \bibinfo{person}{Aida Nematzadeh}, \bibinfo{person}{Elena Gribovskaya},
  \bibinfo{person}{Domenic Donato}, \bibinfo{person}{Angeliki Lazaridou},
  \bibinfo{person}{Arthur Mensch}, \bibinfo{person}{Jean-Baptiste Lespiau},
  \bibinfo{person}{Maria Tsimpoukelli}, \bibinfo{person}{N.~K. Grigorev},
  \bibinfo{person}{Doug Fritz}, \bibinfo{person}{Thibault Sottiaux},
  \bibinfo{person}{Mantas Pajarskas}, \bibinfo{person}{Tobias Pohlen},
  \bibinfo{person}{Zhitao Gong}, \bibinfo{person}{Daniel Toyama},
  \bibinfo{person}{Cyprien de Masson~d'Autume}, \bibinfo{person}{Yujia Li},
  \bibinfo{person}{Tayfun Terzi}, \bibinfo{person}{Vladimir Mikulik},
  \bibinfo{person}{Igor Babuschkin}, \bibinfo{person}{Aidan Clark},
  \bibinfo{person}{Diego de Las~Casas}, \bibinfo{person}{Aurelia Guy},
  \bibinfo{person}{Chris Jones}, \bibinfo{person}{James Bradbury},
  \bibinfo{person}{Matthew~G. Johnson}, \bibinfo{person}{Blake~A. Hechtman},
  \bibinfo{person}{Laura Weidinger}, \bibinfo{person}{Iason Gabriel},
  \bibinfo{person}{William~S. Isaac}, \bibinfo{person}{Edward Lockhart},
  \bibinfo{person}{Simon Osindero}, \bibinfo{person}{Laura Rimell},
  \bibinfo{person}{Chris Dyer}, \bibinfo{person}{Oriol Vinyals},
  \bibinfo{person}{Kareem~W. Ayoub}, \bibinfo{person}{Jeff Stanway},
  \bibinfo{person}{L.~L. Bennett}, \bibinfo{person}{Demis Hassabis},
  \bibinfo{person}{Koray Kavukcuoglu}, {and} \bibinfo{person}{Geoffrey
  Irving}.} \bibinfo{year}{2021}\natexlab{}.
\newblock \showarticletitle{Scaling Language Models: Methods, Analysis \&
  Insights from Training Gopher}.
\newblock \bibinfo{journal}{\emph{ArXiv}}  \bibinfo{volume}{abs/2112.11446}
  (\bibinfo{year}{2021}).
\newblock


\bibitem[Raffel et~al\mbox{.}(2019)]%
        {Raffel2019ExploringTL}
\bibfield{author}{\bibinfo{person}{Colin Raffel}, \bibinfo{person}{Noam~M.
  Shazeer}, \bibinfo{person}{Adam Roberts}, \bibinfo{person}{Katherine Lee},
  \bibinfo{person}{Sharan Narang}, \bibinfo{person}{Michael Matena},
  \bibinfo{person}{Yanqi Zhou}, \bibinfo{person}{Wei Li}, {and}
  \bibinfo{person}{Peter~J. Liu}.} \bibinfo{year}{2019}\natexlab{}.
\newblock \showarticletitle{Exploring the Limits of Transfer Learning with a
  Unified Text-to-Text Transformer}.
\newblock \bibinfo{journal}{\emph{ArXiv}}  \bibinfo{volume}{abs/1910.10683}
  (\bibinfo{year}{2019}).
\newblock


\bibitem[Rombach et~al\mbox{.}(2021)]%
        {Rombach2021HighResolutionIS}
\bibfield{author}{\bibinfo{person}{Robin Rombach}, \bibinfo{person}{A.
  Blattmann}, \bibinfo{person}{Dominik Lorenz}, \bibinfo{person}{Patrick
  Esser}, {and} \bibinfo{person}{Bj{\"o}rn Ommer}.}
  \bibinfo{year}{2021}\natexlab{}.
\newblock \showarticletitle{High-Resolution Image Synthesis with Latent
  Diffusion Models}.
\newblock \bibinfo{journal}{\emph{2022 IEEE/CVF Conference on Computer Vision
  and Pattern Recognition (CVPR)}} (\bibinfo{year}{2021}),
  \bibinfo{pages}{10674--10685}.
\newblock


\bibitem[Rombach et~al\mbox{.}(2022)]%
        {Rombach_2022_CVPR}
\bibfield{author}{\bibinfo{person}{Robin Rombach}, \bibinfo{person}{Andreas
  Blattmann}, \bibinfo{person}{Dominik Lorenz}, \bibinfo{person}{Patrick
  Esser}, {and} \bibinfo{person}{Bj\"orn Ommer}.}
  \bibinfo{year}{2022}\natexlab{}.
\newblock \showarticletitle{High-Resolution Image Synthesis With Latent
  Diffusion Models}. In \bibinfo{booktitle}{\emph{Proceedings of the IEEE/CVF
  Conference on Computer Vision and Pattern Recognition (CVPR)}}.
  \bibinfo{pages}{10684--10695}.
\newblock


\bibitem[Rosenfeld et~al\mbox{.}(2019)]%
        {Rosenfeld2019ACP}
\bibfield{author}{\bibinfo{person}{Jonathan~S. Rosenfeld},
  \bibinfo{person}{Amir Rosenfeld}, \bibinfo{person}{Yonatan Belinkov}, {and}
  \bibinfo{person}{Nir Shavit}.} \bibinfo{year}{2019}\natexlab{}.
\newblock \showarticletitle{A Constructive Prediction of the Generalization
  Error Across Scales}.
\newblock \bibinfo{journal}{\emph{ArXiv}}  \bibinfo{volume}{abs/1909.12673}
  (\bibinfo{year}{2019}).
\newblock


\bibitem[Ruiz et~al\mbox{.}(2022)]%
        {Ruiz2022DreamBoothFT}
\bibfield{author}{\bibinfo{person}{Nataniel Ruiz}, \bibinfo{person}{Yuanzhen
  Li}, \bibinfo{person}{Varun Jampani}, \bibinfo{person}{Yael Pritch},
  \bibinfo{person}{Michael Rubinstein}, {and} \bibinfo{person}{Kfir Aberman}.}
  \bibinfo{year}{2022}\natexlab{}.
\newblock \showarticletitle{DreamBooth: Fine Tuning Text-to-Image Diffusion
  Models for Subject-Driven Generation}.
\newblock \bibinfo{journal}{\emph{ArXiv}}  \bibinfo{volume}{abs/2208.12242}
  (\bibinfo{year}{2022}).
\newblock


\bibitem[Saharia et~al\mbox{.}(2021)]%
        {Saharia2021ImageSV}
\bibfield{author}{\bibinfo{person}{Chitwan Saharia}, \bibinfo{person}{Jonathan
  Ho}, \bibinfo{person}{William Chan}, \bibinfo{person}{Tim Salimans},
  \bibinfo{person}{David~J. Fleet}, {and} \bibinfo{person}{Mohammad Norouzi}.}
  \bibinfo{year}{2021}\natexlab{}.
\newblock \showarticletitle{Image Super-Resolution via Iterative Refinement}.
\newblock \bibinfo{journal}{\emph{IEEE Transactions on Pattern Analysis and
  Machine Intelligence}}  \bibinfo{volume}{45} (\bibinfo{year}{2021}),
  \bibinfo{pages}{4713--4726}.
\newblock


\bibitem[Shoeybi et~al\mbox{.}(2019)]%
        {Shoeybi2019MegatronLMTM}
\bibfield{author}{\bibinfo{person}{Mohammad Shoeybi}, \bibinfo{person}{Mostofa
  Patwary}, \bibinfo{person}{Raul Puri}, \bibinfo{person}{Patrick LeGresley},
  \bibinfo{person}{Jared Casper}, {and} \bibinfo{person}{Bryan Catanzaro}.}
  \bibinfo{year}{2019}\natexlab{}.
\newblock \showarticletitle{Megatron-LM: Training Multi-Billion Parameter
  Language Models Using Model Parallelism}.
\newblock \bibinfo{journal}{\emph{ArXiv}}  \bibinfo{volume}{abs/1909.08053}
  (\bibinfo{year}{2019}).
\newblock


\bibitem[Smith et~al\mbox{.}(2022)]%
        {Smith2022UsingDA}
\bibfield{author}{\bibinfo{person}{Shaden Smith}, \bibinfo{person}{Mostofa
  Patwary}, \bibinfo{person}{Brandon Norick}, \bibinfo{person}{Patrick
  LeGresley}, \bibinfo{person}{Samyam Rajbhandari}, \bibinfo{person}{Jared
  Casper}, \bibinfo{person}{Zhun Liu}, \bibinfo{person}{Shrimai Prabhumoye},
  \bibinfo{person}{George Zerveas}, \bibinfo{person}{Vijay~Anand Korthikanti},
  \bibinfo{person}{Elton Zhang}, \bibinfo{person}{Rewon Child},
  \bibinfo{person}{Reza~Yazdani Aminabadi}, \bibinfo{person}{Julie Bernauer},
  \bibinfo{person}{Xia Song}, \bibinfo{person}{Mohammad Shoeybi},
  \bibinfo{person}{Yuxiong He}, \bibinfo{person}{Michael Houston},
  \bibinfo{person}{Saurabh Tiwary}, {and} \bibinfo{person}{Bryan Catanzaro}.}
  \bibinfo{year}{2022}\natexlab{}.
\newblock \showarticletitle{Using DeepSpeed and Megatron to Train
  Megatron-Turing NLG 530B, A Large-Scale Generative Language Model}.
\newblock \bibinfo{journal}{\emph{ArXiv}}  \bibinfo{volume}{abs/2201.11990}
  (\bibinfo{year}{2022}).
\newblock


\bibitem[Touvron et~al\mbox{.}(2023)]%
        {touvron2023llama}
\bibfield{author}{\bibinfo{person}{Hugo Touvron}, \bibinfo{person}{Thibaut
  Lavril}, \bibinfo{person}{Gautier Izacard}, \bibinfo{person}{Xavier
  Martinet}, \bibinfo{person}{Marie-Anne Lachaux},
  \bibinfo{person}{Timoth{\'e}e Lacroix}, \bibinfo{person}{Baptiste
  Rozi{\`e}re}, \bibinfo{person}{Naman Goyal}, \bibinfo{person}{Eric Hambro},
  \bibinfo{person}{Faisal Azhar}, {et~al\mbox{.}}}
  \bibinfo{year}{2023}\natexlab{}.
\newblock \showarticletitle{Llama: Open and efficient foundation language
  models}.
\newblock \bibinfo{journal}{\emph{arXiv preprint arXiv:2302.13971}}
  (\bibinfo{year}{2023}).
\newblock


\bibitem[Valevski et~al\mbox{.}(2022)]%
        {Valevski2022UniTuneTI}
\bibfield{author}{\bibinfo{person}{Dani Valevski}, \bibinfo{person}{Matan
  Kalman}, \bibinfo{person}{Y. Matias}, {and} \bibinfo{person}{Yaniv
  Leviathan}.} \bibinfo{year}{2022}\natexlab{}.
\newblock \showarticletitle{UniTune: Text-Driven Image Editing by Fine Tuning
  an Image Generation Model on a Single Image}.
\newblock \bibinfo{journal}{\emph{ArXiv}}  \bibinfo{volume}{abs/2210.09477}
  (\bibinfo{year}{2022}).
\newblock


\bibitem[Vaswani et~al\mbox{.}(2017)]%
        {vaswani2017attention}
\bibfield{author}{\bibinfo{person}{Ashish Vaswani}, \bibinfo{person}{Noam
  Shazeer}, \bibinfo{person}{Niki Parmar}, \bibinfo{person}{Jakob Uszkoreit},
  \bibinfo{person}{Llion Jones}, \bibinfo{person}{Aidan~N Gomez},
  \bibinfo{person}{{\L}ukasz Kaiser}, {and} \bibinfo{person}{Illia
  Polosukhin}.} \bibinfo{year}{2017}\natexlab{}.
\newblock \showarticletitle{Attention is all you need}.
\newblock \bibinfo{journal}{\emph{Advances in neural information processing
  systems}}  \bibinfo{volume}{30} (\bibinfo{year}{2017}).
\newblock


\bibitem[Wang et~al\mbox{.}(2022)]%
        {wang2022exploring}
\bibfield{author}{\bibinfo{person}{Jianyi Wang}, \bibinfo{person}{Kelvin~CK
  Chan}, {and} \bibinfo{person}{Chen~Change Loy}.}
  \bibinfo{year}{2022}\natexlab{}.
\newblock \showarticletitle{Exploring CLIP for Assessing the Look and Feel of
  Images}.
\newblock \bibinfo{journal}{\emph{arXiv preprint arXiv:2207.12396}}
  (\bibinfo{year}{2022}).
\newblock


\bibitem[Wei et~al\mbox{.}(2022)]%
        {Wei2022EmergentAO}
\bibfield{author}{\bibinfo{person}{Jason Wei}, \bibinfo{person}{Yi Tay},
  \bibinfo{person}{Rishi Bommasani}, \bibinfo{person}{Colin Raffel},
  \bibinfo{person}{Barret Zoph}, \bibinfo{person}{Sebastian Borgeaud},
  \bibinfo{person}{Dani Yogatama}, \bibinfo{person}{Maarten Bosma},
  \bibinfo{person}{Denny Zhou}, \bibinfo{person}{Donald Metzler},
  \bibinfo{person}{Ed~Huai hsin Chi}, \bibinfo{person}{Tatsunori Hashimoto},
  \bibinfo{person}{Oriol Vinyals}, \bibinfo{person}{Percy Liang},
  \bibinfo{person}{Jeff Dean}, {and} \bibinfo{person}{William Fedus}.}
  \bibinfo{year}{2022}\natexlab{}.
\newblock \showarticletitle{Emergent Abilities of Large Language Models}.
\newblock \bibinfo{journal}{\emph{ArXiv}}  \bibinfo{volume}{abs/2206.07682}
  (\bibinfo{year}{2022}).
\newblock


\bibitem[Yang et~al\mbox{.}(2022b)]%
        {Yang2022DiffusionMA}
\bibfield{author}{\bibinfo{person}{Ling Yang}, \bibinfo{person}{Zhilong Zhang},
  \bibinfo{person}{Shenda Hong}, \bibinfo{person}{Runsheng Xu},
  \bibinfo{person}{Yue Zhao}, \bibinfo{person}{Yingxia Shao},
  \bibinfo{person}{Wentao Zhang}, \bibinfo{person}{Ming-Hsuan Yang}, {and}
  \bibinfo{person}{Bin Cui}.} \bibinfo{year}{2022}\natexlab{b}.
\newblock \showarticletitle{Diffusion Models: A Comprehensive Survey of Methods
  and Applications}.
\newblock \bibinfo{journal}{\emph{ArXiv}}  \bibinfo{volume}{abs/2209.00796}
  (\bibinfo{year}{2022}).
\newblock


\bibitem[Yang et~al\mbox{.}(2022a)]%
        {Yang2022DiffusionPM}
\bibfield{author}{\bibinfo{person}{Ruihan Yang}, \bibinfo{person}{Prakhar
  Srivastava}, {and} \bibinfo{person}{Stephan Mandt}.}
  \bibinfo{year}{2022}\natexlab{a}.
\newblock \showarticletitle{Diffusion Probabilistic Modeling for Video
  Generation}.
\newblock \bibinfo{journal}{\emph{ArXiv}}  \bibinfo{volume}{abs/2203.09481}
  (\bibinfo{year}{2022}).
\newblock


\bibitem[Zeng et~al\mbox{.}(2022)]%
        {Zeng2022GLM130BAO}
\bibfield{author}{\bibinfo{person}{Aohan Zeng}, \bibinfo{person}{Xiao Liu},
  \bibinfo{person}{Zhengxiao Du}, \bibinfo{person}{Zihan Wang},
  \bibinfo{person}{Hanyu Lai}, \bibinfo{person}{Ming Ding},
  \bibinfo{person}{Zhuoyi Yang}, \bibinfo{person}{Yifan Xu},
  \bibinfo{person}{Wendi Zheng}, \bibinfo{person}{Xiao Xia},
  \bibinfo{person}{Weng~Lam Tam}, \bibinfo{person}{Zixuan Ma},
  \bibinfo{person}{Yufei Xue}, \bibinfo{person}{Jidong Zhai},
  \bibinfo{person}{Wenguang Chen}, \bibinfo{person}{P. Zhang},
  \bibinfo{person}{Yuxiao Dong}, {and} \bibinfo{person}{Jie Tang}.}
  \bibinfo{year}{2022}\natexlab{}.
\newblock \showarticletitle{GLM-130B: An Open Bilingual Pre-trained Model}.
\newblock \bibinfo{journal}{\emph{ArXiv}}  \bibinfo{volume}{abs/2210.02414}
  (\bibinfo{year}{2022}).
\newblock


\bibitem[Zhang et~al\mbox{.}(2023)]%
        {Zhang2023TexttoimageDM}
\bibfield{author}{\bibinfo{person}{Chenshuang Zhang}, \bibinfo{person}{Chaoning
  Zhang}, \bibinfo{person}{Mengchun Zhang}, {and} \bibinfo{person}{In-So
  Kweon}.} \bibinfo{year}{2023}\natexlab{}.
\newblock \showarticletitle{Text-to-image Diffusion Models in Generative AI: A
  Survey}.
\newblock \bibinfo{journal}{\emph{ArXiv}}  \bibinfo{volume}{abs/2303.07909}
  (\bibinfo{year}{2023}).
\newblock


\bibitem[Zhang and Agrawala(2023a)]%
        {Zhang2023AddingCC}
\bibfield{author}{\bibinfo{person}{Lvmin Zhang} {and} \bibinfo{person}{Maneesh
  Agrawala}.} \bibinfo{year}{2023}\natexlab{a}.
\newblock \showarticletitle{Adding Conditional Control to Text-to-Image
  Diffusion Models}.
\newblock \bibinfo{journal}{\emph{ArXiv}}  \bibinfo{volume}{abs/2302.05543}
  (\bibinfo{year}{2023}).
\newblock


\bibitem[Zhang and Agrawala(2023b)]%
        {zhang2023adding}
\bibfield{author}{\bibinfo{person}{Lvmin Zhang} {and} \bibinfo{person}{Maneesh
  Agrawala}.} \bibinfo{year}{2023}\natexlab{b}.
\newblock \bibinfo{title}{Adding Conditional Control to Text-to-Image Diffusion
  Models}.
\newblock
\newblock
\showeprint[arxiv]{2302.05543}~[cs.CV]


\bibitem[Zhang et~al\mbox{.}(2022a)]%
        {Zhang2022MotionDiffuseTH}
\bibfield{author}{\bibinfo{person}{Mingyuan Zhang}, \bibinfo{person}{Zhongang
  Cai}, \bibinfo{person}{Liang Pan}, \bibinfo{person}{Fangzhou Hong},
  \bibinfo{person}{Xinying Guo}, \bibinfo{person}{Lei Yang}, {and}
  \bibinfo{person}{Ziwei Liu}.} \bibinfo{year}{2022}\natexlab{a}.
\newblock \showarticletitle{MotionDiffuse: Text-Driven Human Motion Generation
  with Diffusion Model}.
\newblock \bibinfo{journal}{\emph{ArXiv}}  \bibinfo{volume}{abs/2208.15001}
  (\bibinfo{year}{2022}).
\newblock


\bibitem[Zhang et~al\mbox{.}(2022b)]%
        {Zhang2022OPTOP}
\bibfield{author}{\bibinfo{person}{Susan Zhang}, \bibinfo{person}{Stephen
  Roller}, \bibinfo{person}{Naman Goyal}, \bibinfo{person}{Mikel Artetxe},
  \bibinfo{person}{Moya Chen}, \bibinfo{person}{Shuohui Chen},
  \bibinfo{person}{Christopher Dewan}, \bibinfo{person}{Mona Diab},
  \bibinfo{person}{Xian Li}, \bibinfo{person}{Xi~Victoria Lin},
  \bibinfo{person}{Todor Mihaylov}, \bibinfo{person}{Myle Ott},
  \bibinfo{person}{Sam Shleifer}, \bibinfo{person}{Kurt Shuster},
  \bibinfo{person}{Daniel Simig}, \bibinfo{person}{Punit~Singh Koura},
  \bibinfo{person}{Anjali Sridhar}, \bibinfo{person}{Tianlu Wang}, {and}
  \bibinfo{person}{Luke Zettlemoyer}.} \bibinfo{year}{2022}\natexlab{b}.
\newblock \showarticletitle{OPT: Open Pre-trained Transformer Language Models}.
\newblock \bibinfo{journal}{\emph{ArXiv}}  \bibinfo{volume}{abs/2205.01068}
  (\bibinfo{year}{2022}).
\newblock


\bibitem[Zhong et~al\mbox{.}(2023)]%
        {Zhong2023SPEMSP}
\bibfield{author}{\bibinfo{person}{Shan Zhong}, \bibinfo{person}{Wushao Wen},
  {and} \bibinfo{person}{Jinghui Qin}.} \bibinfo{year}{2023}\natexlab{}.
\newblock \showarticletitle{SPEM: Self-adaptive Pooling Enhanced Attention
  Module for Image Recognition}. In \bibinfo{booktitle}{\emph{Conference on
  Multimedia Modeling}}.
\newblock


\end{thebibliography}

\clearpage

\appendix

\section{Supplemental Dataset Information}

\subsection{Pre-trained Models}

\noindent\textbf{BLIP. }
We utilize the BLIP (Bootstrapping Language-Image Pre-training)~\cite{li2022blip,guo2022images} model to generate simple narrative prompts of images for SURD. Specifically, we employ the BLIP caption base model, which has been fine-tuned on the MSCOCO~\cite{Lin2014MicrosoftCC} captioning dataset, using the load function provided in the official documentation~\footnote{https://github.com/salesforce/LAVIS}.

\noindent\textbf{CLIP. }
We utilize CLIP to ensure the correctness of both simple narrative prompts and complex keyword-based prompts. Specifically, we designed a data cleaning process, which is briefly described in Section 3.1 of the main text.
We leverage the semantic similarity between images and prompts by asking CLIP to classify between simple and complex prompts, where the goal is to select the prompts that best match the semantics of the images. Typically, complex prompts contain semantically irrelevant information, such as image quality descriptions, and therefore, semantically correct simple prompts generally achieve higher CLIP scores than complex prompts. We retain a sample if the CLIP score of the corresponding simple prompt is not lower than that of the complex prompt. We use the publicly available pre-trained CLIP model, which has a ViT-B/32 architecture, and load it using the function provided in the official documentation~\footnote{https://github.com/openai/CLIP}.

\noindent\textbf{LLMs. }
In this paper, we utilize LLaMA~\cite{touvron2023llama}, a collection of foundation language models ranging from 7B to 65B parameters, as knowledge distillation for large language models (LLMs). Specifically, we save the vector representations of simple prompts in LLMs, which serve as the text understanding to finetune diffusion models. The details of the LLMs used in our experiments, including the number of parameters, vector dimensions, and model structures, are shown in Table~\ref{tab:llms}.

\begin{table}[htbp]
  \vspace{-0.3cm}
  \centering
  \caption{Model sizes and architectures of LLMs used in the main text. } 
  \vspace{-0.3cm}
    \resizebox*{0.8\linewidth}{!}{
    \begin{tabular}{c|cccc}
    \toprule
    LLM  & params & dimension & n heads & n layers \\
    \midrule
    7B    & 6.7B  & 4096  & 32    & 32 \\
    13B   & 13.0B & 5120  & 40    & 40 \\
    33B   & 32.5B & 6656  & 52    & 60 \\
    \bottomrule
    \end{tabular}%
    }
  \label{tab:llms}%
  \vspace{-0.5cm}
\end{table}%

\subsection{Impact and Ethics}

\noindent\textbf{Impact and Usage. }
Improving the SUR ability of diffusion models is an important issue that has received limited attention in the research community. In this paper, we approach this problem from a novel perspective by constructing a semantically correct dataset, SURD, and using knowledge distillation to transfer semantic knowledge from complex prompts and LLM. SURD can not only be used to finetune diffusion models for solving SUR problems but can also be directly used as a training dataset for diffusion models due to its ensured semantic correctness.
 
\noindent\textbf{Social Ethics. }Unlike many multimodal datasets in the natural domain, SURD is entirely built on data generated by DNNs. As a result, it is less likely to be used in surveillance systems that could potentially violate people's privacy. Moreover, during the data cleaning, a manual inspection stage ensures that SURD does not contain any sensitive personal information, such as gender and race, nor does it include data that could exacerbate biases towards underrepresented communities. Therefore, upon careful examination of our dataset, we believe that it is unlikely to be used to directly harm individuals.

\begin{figure}[htp]
 \centering
 \includegraphics[width=\linewidth]{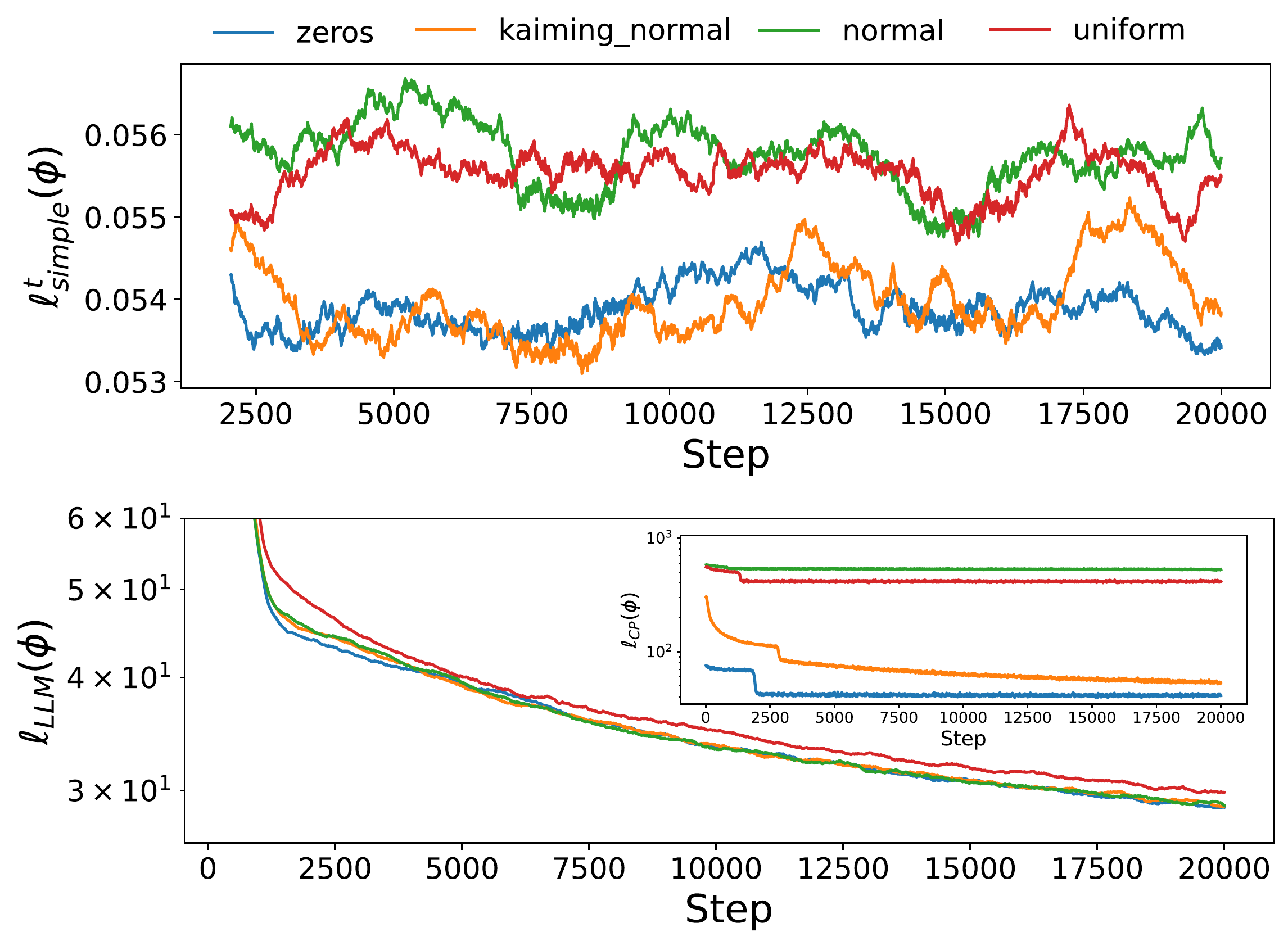}
 \caption{Loss value during the training of SUR-adapter with different initializations. The mathematical symbols correspond to Eq.(10).}
 \label{fig:loss}
 \vspace{-0.3cm}
\end{figure}

\begin{table*}[htbp]
  \vspace{-0.3cm}
  \centering
  \caption{Evaluation of semantic accuracy (Acc.) in images generated by simple prompts using diffusion models. The simple prompts consisted of three types of sentences, including "counting", "color", and "action". Each prompt generated 130 images, and the images were manually checked for semantic accuracy. }
  \vspace{-0.3cm}
  \resizebox*{0.8\linewidth}{!}{
    \begin{tabular}{llcc}
    \toprule
    Type & Prompt & Accuracy & Accuracy (Ours) \\
    \midrule
    \multirow{3}[2]{*}{Counting} & Four freshly baked pies. & 63.08\% & 73.85\% \\
          & Six colorful hot air balloons floating over a picturesque landscape. & 8.46\% & 41.54\% \\
          & Seven vintage glass bottles. & 0.00\% & 36.92\% \\
    \midrule
    \multirow{3}[2]{*}{Color} & A vibrant red sports car speeding down a winding road. & 86.15\% & 93.85\% \\
          & The blue glass containing red juice. & 17.69\% & 20.00\% \\
          & A couple wearing blue and yellow solid color clothes respectively. & 0.00\% & 6.92\% \\
    \midrule
    \multirow{3}[2]{*}{Action} & Someone shooting a basketball on the sports field. & 41.54\% & 56.92\% \\
          & Giraffes eating trees. & 25.38\% & 50.77\% \\
          & A chef tossing a pizza dough in the air in a kitchen. & 15.38\% & 32.31\% \\
    \bottomrule
    \end{tabular}%
  \label{tab:intro_sura}%
  }
  \vspace{-0.3cm}
\end{table*}%

\begin{table*}[htbp]
  \centering
  \caption{Examples of testing prompts. }
  \vspace{-0.3cm}
  \resizebox*{0.7\linewidth}{!}{
    \begin{tabular}{ll}
    \toprule
    Type  & Prompt \\
    \midrule
    \multirow{3}[2]{*}{Action} 
          & A gymnast performing a balance beam routine with graceful flips and twists. \\
          & A skateboarder doing a kickflip over a set of stairs. \\
          & A diver swimming underwater with colorful fish and coral all around him. \\
    \midrule
    \multirow{3}[2]{*}{Color} 
          & A golden sun setting over a calm ocean, with orange and pink hues appearing in the sky. \\
          & A tranquil scene of a meadow filled with wildflowers in shades of purple, pink, and yellow. \\
          & A funky and retro diner with a color scheme of bright pink, teal, and silver. \\
    \midrule
    \multirow{3}[2]{*}{Counting} 
          & A set of four antique teacups and saucers with intricate floral designs. \\
          & Five different types of fresh fruit cut into slices and arranged on a platter. \\
          & Seven colorful beach umbrellas on a sandy beach. \\
    \bottomrule
    \end{tabular}%
    }
  \label{tab:test_prompt}%
  \vspace{-0.3cm}
\end{table*}%

\section{Supplemental Experiments}

\subsection{Supplemental Implementation details}
In our study, we validate the universality of SUR-adapter with two pre-trained diffusion models, three LLMs with different parameters, and various controlled methods. 
Unless otherwise specified, we follow the settings of~\cite{Rombach_2022_CVPR,touvron2023llama,zhang2023adding,bar2023multidiffusion,hong2022improving}. 
Specifically, all models are trained on one Nvidia RTX 3090 GPU, with step set to 5000, batch size set to 16, and resolution set to 512. During training, we apply mixed precision and standard data augmentation techniques such as normalization, center cropping, and horizontal flipping. The learning rate and hyper-parameters in Eq.(7) and Eq.(10) are set to 1e-5.

All control methods utilize the default settings of diffusers~\footnote{https://github.com/huggingface/diffusers}. Besides, we manually curated a set of images that satisfy the semantic requirements. These images serve as conditional inputs for ControlNet (canny) and ControlNet (seg). The setting of MultiDiffusion is that the pretrained model for DM (1.5) uses the schedulers of DM (cartoon), and vice versa, the pretrained model for DM (cartoon) uses the schedulers of DM (1.5).

\subsection{The Initiation of SUR-adapter}
As shown in Fig. 5, we use a fully connected network to connect the adapter and the backbone. To ensure stable training of the adapter, we initialize the FCN with 0, following some well-known adapter-related works \cite{hu2021lora,Zhang2023AddingCC}. Additionally, as shown in Fig. \ref{fig:loss}, we also demonstrate the impact of different initialization methods on the loss of SUR-adapter. We observe that different initializations have little impact on $\ell_{\text{LLM}}(\phi)$ in Eq.(10), but have a significant effect on the training of $\ell_{\text{CP}}(\phi)$ and the diffusion model, which is consistent with existing works \cite{hu2021lora,Zhang2023AddingCC}.

\subsection{Accuracy of SUR-adapter in Table 1}

We have provided additional information on the semantic accuracy of SUR-adapter prompts in Table~\ref{tab:intro_sura}, which supplements the prompt examples shown in Table 1 of the Introduction.

\subsection{User Preference Study}
\begin{figure}
  \centering
  \includegraphics[width=\linewidth]{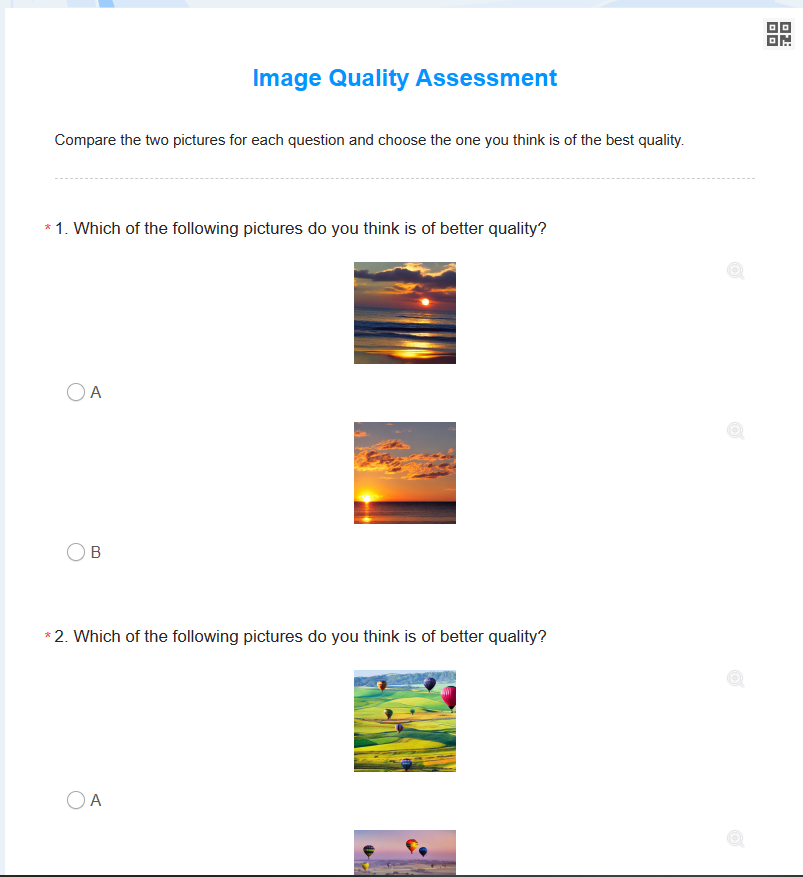}
  \caption{Title, description, and some questions of the user preference study.}
  \label{fig:ques}
\end{figure}

In this paper, there are two metrics that require manual judgment. One is the semantic accuracy of the generated images (action, color, counting), which is an objective metric. Therefore, it can be easily assessed and counted by the authors. The other metric that requires manual judgment is user preference, as shown in Table 3. This metric is subjective. To gather data for this metric, we collected a total of 89 valid questionnaires (an example of the questionnaire is provided in Fig.~\ref{fig:ques}). We randomly presented images generated by our method and baselines to the participants and asked them to select a picture that they deemed of better quality based on the question, "Which of the following pictures do you think is of better quality?" Finally, based on the 89 questionnaires, we compiled and analyzed the data.

\subsection{Testing Prompts}
To evaluate Semantic Understanding and Reasoning (SUR), we have divided the semantics into three main types, namely Action, Color, and Counting, with each type having fifteen prompts whose examples are shown in Table~\ref{tab:test_prompt}. For each prompt, we generate ten images during testing.

\end{document}